\documentclass[lettersize,journal]{IEEEtran}
\usepackage{amsmath,amsfonts}
\usepackage{amssymb}
\usepackage{algpseudocode}
\usepackage{algorithm}
\usepackage{array}
\usepackage[caption=false,font=footnotesize,labelfont=sf,textfont=sf]{subfig}
\usepackage{enumitem}
\usepackage{textcomp}
\usepackage{booktabs}
\usepackage{multirow}
\usepackage{multicol}
\usepackage{stfloats}
\usepackage{url}
\usepackage{verbatim}
\usepackage{graphicx}
\usepackage{tabularx}
\usepackage{ragged2e}
\usepackage{xspace}
\usepackage{cite}
\usepackage{comment}
\usepackage{listings}
\usepackage{hyperref}
\usepackage[normalem]{ulem}
\usepackage[framemethod=default]{mdframed}
\newmdenv[backgroundcolor=orange, linecolor=orange]{highlightedbox}
\usepackage{soul}
\usepackage[dvipsnames]{xcolor}
\usepackage{pifont}

\newcommand{\Name}{\texttt{TMO}\xspace}
\newcommand{\DatasetName}{\textsc{M4A1}\xspace}

\usepackage{colortbl}
\colorlet{Yes}{blue!15}
\colorlet{No}{red!15}

\def\BibTeX{{\rm B\kern-.05em{\sc i\kern-.025em b}\kern-.08em
    T\kern-.1667em\lower.7ex\hbox{E}\kern-.125emX}}
\makeatletter
\def\subsubsection{\@startsection{subsubsection}{3}{1.25\parindent}{0.1ex plus 0.1ex minus 0.1ex}
{0.1ex}{\normalfont\normalsize\itshape}}
\makeatother

\begin{document}

\title{
Device-Cloud Collaborative LLM Inference with Multi-Modal, Multi-Task, Multi-Turn Conversations
}

\author{Liangqi Yuan,~\IEEEmembership{Graduate Student Member,~IEEE,}
        Dong-Jun Han,~\IEEEmembership{Member,~IEEE,}
        Shiqiang Wang,~\IEEEmembership{Fellow,~IEEE,}
        and Christopher G. Brinton,~\IEEEmembership{Senior Member,~IEEE}

\thanks{This work was supported in part by the National Science Foundation (NSF) under grant CPS-2313109, by the Office of Naval Research (ONR) under grant N00014-23-C-1016, by the Air Force Office of Scientific Research (AFOSR) under grant FA9550-24-1-0083, by the Institute of Information \& communications Technology Planning \& Evaluation (IITP) from the Korea government (MSIT) (No. RS-2024-00457882, AI Research Hub Project), and by NVIDIA's Academic Grant Program.}
\thanks{An earlier version of this paper was presented at the Twenty-sixth International Symposium on Theory, Algorithmic Foundations, and Protocol Design for Mobile Networks and Mobile Computing (MobiHoc 2025)~\cite{yuan2025local}, where it received the Best Paper Award Runner-up.}
\thanks{L. Yuan and C. G. Brinton are with the School of Electrical and Computer Engineering, Purdue University, West Lafayette, IN 47907, USA.  E-mail: liangqiy@purdue.edu; cgb@purdue.edu}
\thanks{D.-J. Han is with the Department of Computer Science and Engineering, Yonsei University, Seoul, South Korea. E-mail: djh@yonsei.ac.kr}
\thanks{S. Wang is with the Department of Computer Science, University of Exeter, EX4 4RN, UK. E-mail: shiqiang.wang@ieee.org}
}

\maketitle

\begin{abstract}
Compared to traditional machine learning models, recent large language models (LLMs) can exhibit multi-task-solving capabilities through multi-modal data sources and multi-turn conversations. These unique characteristics of LLMs, together with their large model size, make their deployment more challenging. Specifically, (i) deploying LLMs on devices faces computational, memory, and energy resource issues, while (ii) deploying them in the cloud cannot guarantee real-time service and incurs communication/usage costs.
In this paper, we design \Name, a device-cloud LLM inference system with Three-M Offloading: Multi-modal, Multi-task, and Multi-turn conversation. \Name incorporates (i) a lightweight on-device LLM that can process simple tasks at high speed and (ii) a large-scale cloud LLM that can handle multi-modal data sources. We develop a resource-constrained reinforcement learning (RCRL) strategy for \Name that optimizes the inference location (i.e., device vs. cloud) and multi-modal data sources to use for each task in multi-turn conversations, aiming to maximize the long-term reward (response quality, latency, and usage cost) while adhering to resource constraints. 
We also contribute \DatasetName, a new dataset we curated across multiple modalities, tasks, conversation turns, and LLM configurations, enabling evaluation of offloading decisions. We demonstrate the effectiveness of \Name compared to several exploration-decision and LLM-as-Router baselines, showing significant improvements in latency, cost, and response quality. Our code and dataset are available at \url{https://github.com/liangqiyuan/TMO}.
\end{abstract}

\begin{IEEEkeywords}
Large Language Model, Reinforcement Learning, Multi-modal, Collaborative Inference, Resource Constraint.
\end{IEEEkeywords}

\section{Introduction}
\label{Sec. Introduction}

Large language models (LLMs) have demonstrated remarkable general-purpose task-solving capabilities across various fields and have been gradually incorporated into daily life \cite{achiam2023gpt, zhao2023survey, yuan2025llmap, yuan2025next}. The development of LLMs is thriving, yet deploying LLMs on devices presents practical challenges, including computational, memory, and energy resource limitations \cite{su2024titanic, wang2024resource, yuan2026paac}. These issues hinder the deployment of large-scale LLMs on users' devices. Industry giants have developed strategies that either (i) involve lightweight versions of LLMs, such as Google's Gemma and Meta's LLaMA, which offer solutions with varying numbers of parameters (typically 3B, 8B, 13B, etc.) for fast and cost-efficient response, or (ii) place LLMs in the cloud, where users interact with them via the Internet, as seen with the ChatGPT application on mobile phones. The former often suffers from inferior inference performance, while the latter requires substantial network bandwidth and incurs high costs from service providers \cite{yuan2026large}. This creates a pressing need to build systems that combine the advantages of both strategies, i.e., enhancing response quality while ensuring lower latency and usage costs.

\begin{figure}[t]
    \centering
    \includegraphics[width=1\linewidth]{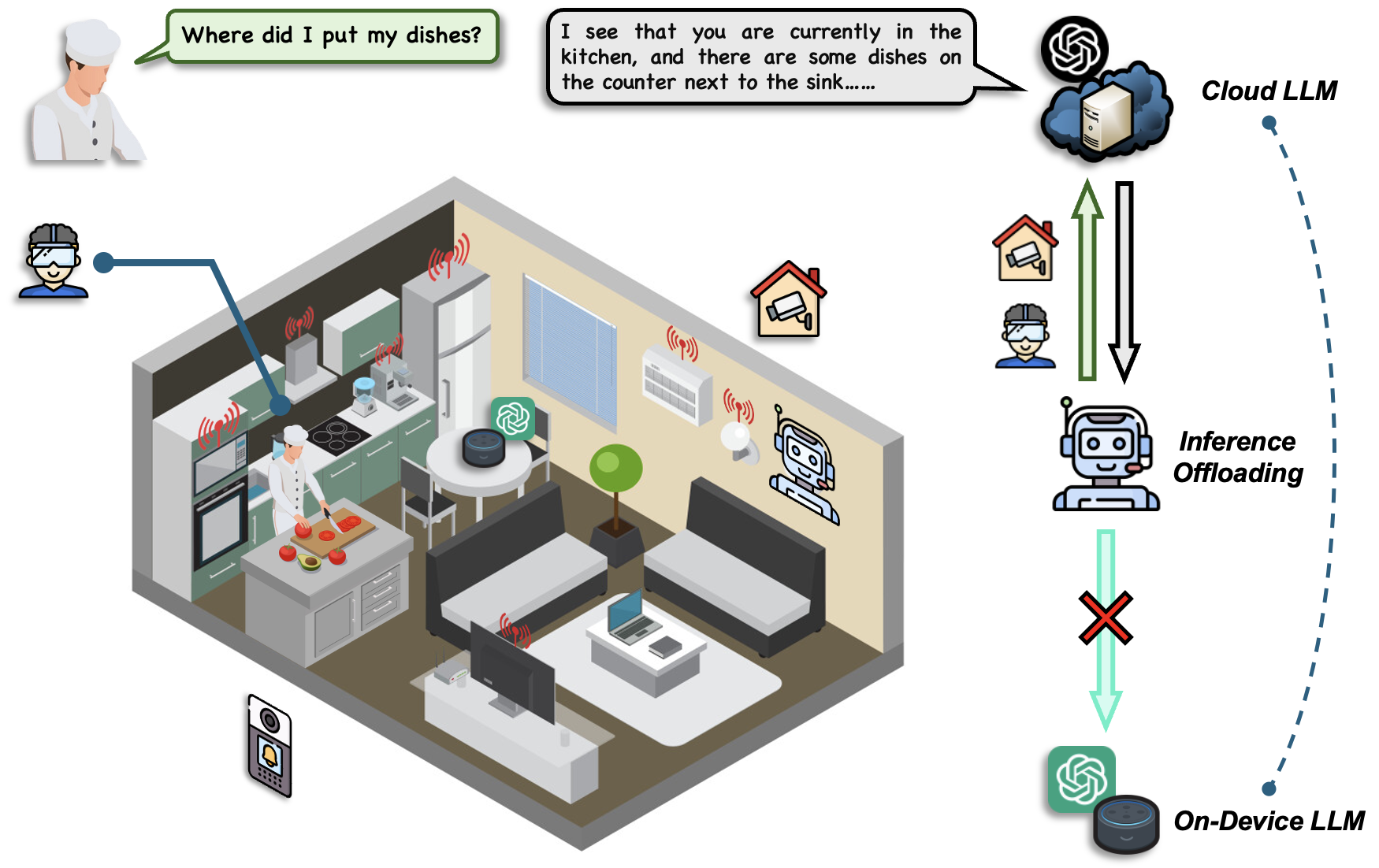}
    \caption{\textbf{Application Scenario for the \texttt{TMO} System in Kitchen Activity Assistance.} Given a user prompt, the decision engine selects between the on-device and cloud LLMs and chooses the relevant data modalities based on the current task and conversation history. For the query ``Where did I put my dishes?'', first-person and overhead views are uploaded to the cloud LLM, which the text-only on-device LLM alone could not have solved.}
    \label{Fig. TMO}
\vspace{-4mm}
\end{figure}

\subsection{Research Questions}

Optimizing LLMs over networks introduces unique challenges beyond traditional machine learning models, as LLMs operate with (i) multi-modal data sources (e.g., text, images), (ii) multi-task environments (e.g., editing, recommendations), and (iii) multi-turn conversation contexts (i.e., ongoing conversations with users). Although these elements enrich the capabilities of LLMs, they introduce significant challenges, particularly in the inference stage:

\textit{\textbf{1. Determining the optimal inference location:}} Determining whether to process tasks using an on-device LLM or to offload them to a cloud LLM is non-trivial. This decision critically impacts not only the response quality but also the latency and costs. For example, on-device LLM inference may limit response quality due to the lightweight model deployed on user devices, whereas cloud LLM inference, although more powerful and capable of multi-modal processing, can introduce significant communication and computation delay.

\textit{\textbf{2. Understanding relationships between modalities, tasks, and conversation turns:}} There are intricate associations between multi-modal data sources, inference tasks, and their temporal variation across multiple turns in a conversation. Although various types of input data can improve the response quality of LLMs, not all tasks require comprehensive multi-modal data. For example, a message editing task may not require images as inputs, whereas some human activity tasks require various types of image modalities \cite{lee2024gazepointar}. Furthermore, when considering multi-turn conversations, where each turn represents a task in sequential order, earlier uploaded data modalities may provide sustained informational advantages, yet they might necessitate re-uploading if the information becomes outdated.

\textit{\textbf{3. Uncertainty issues in offline training with LLM inference data:}} It is often desirable to train auxiliary models using LLM inference results, e.g., to learn modality-task associations. The computationally expensive nature of LLMs renders it impractical to conduct online training of such models. Consequently, generating datasets for offline training becomes necessary. However, due to inherent uncertainty in LLM inference, which arises from the probabilistic nature of language model outputs \cite{hu2023uncertainty}, such datasets may contain multiple identical samples with varying response scores. This uncertainty substantially complicates the evaluation of LLM outputs, particularly in multi-task or generative scenarios where standard answers or evaluation metrics do not exist.

Motivated by these challenges, we aim to answer the following research questions. \textit{In the context of device-cloud collaborative LLM systems:}
\begin{itemize}[leftmargin=11mm] 
    \item[(RQ1)] \textit{How can we optimally select the appropriate LLM for inference (i.e., on-device LLM vs. multi-modal cloud LLM)?}
    
    \item[(RQ2)] \textit{How can we identify and utilize the most informative and relevant multi-modal data when necessary for tasks in multi-turn conversations, while balancing response quality, latency, and usage cost?}

    \item[(RQ3)] \textit{How can we achieve robust generalization to arbitrary user-specified resource budgets, enabling flexible adaptation during real-world deployment?}
    
    \item[(RQ4)] \textit{How can we effectively address the inherent uncertainty in LLM inference data during offline training?}
\end{itemize}

\subsection{Summary of Contributions}
In this paper, we develop \Name, a device-cloud collaborative LLM inference system shown in Fig. \ref{Fig. TMO}. \Name orchestrates ``Three-M'' Offloading -- multi-modal, multi-task, and multi-turn -- aimed at enhancing response quality and reducing latency and usage costs. We address the above research questions through a resource-constrained reinforcement learning (RCRL) methodology that learns optimal actions (i.e., RQ1, RQ2, and RQ3) to respond to multi-task scenarios in multi-turn conversations. We develop an estimator of LLM inference through a nearest neighbor strategy for effective offline training (i.e., RQ4). Our main contributions are as follows:
\begin{itemize} 
    \item We construct \Name to enable adaptive offloading between on-device and cloud LLM inference across multi-modal, multi-task, multi-turn conversations. This includes formalizing the interaction with both types of LLMs to dynamically adapt with diverse conversational demands \textbf{(Sec. \ref{Sec. Multi-Task Multi-Turn with Multi-Modal Data Sources} and Sec. \ref{Sec. Device-Cloud LLM Inference})}. 
    
    \item We formulate \Name's decision engine in terms of resource-constrained reinforcement learning (RCRL) to maximize cumulative reward across multi-turn conversations by selecting the optimal LLMs and modalities for inference. This approach enables a concrete optimization of trade-offs between response quality, latency, and usage costs while incorporating task-modality associations into the decision-making process \textbf{(Sec. \ref{Sec. Formulation} and Sec. \ref{Sec. MDP Optimization})}. 
    
    \item We devise a nearest neighbor strategy to overcome state-action pair uncertainty in estimating response quality during RCRL training. Our approach addresses two types of uncertainty that manifest in LLM evaluation contexts: \textit{non-deterministic evaluation (NDE)}, where identical state-action pairs can yield varying response quality evaluations, and \textit{out-of-distribution (OOD)}, where the system estimates the response quality for unseen state-action pairs (\textbf{Sec. \ref{Sec. Response Score}}).
    
    \item We curate a new dataset, termed \DatasetName, which contains samples from three multi-modal data sources, four inference tasks, 2-5 conversation turns, and four LLMs, with real-world measurements of latency and usage costs~\textbf{(Sec.~\ref{Sec. Dataset})}. Through extensive evaluation with this dataset, we demonstrate \Name's superior performance in response quality, latency, and usage cost compared to two SOTA exploration-decision and 15 LLM-as-Router baselines~\textbf{(Sec.~\ref{Sec. Experiment})}.
\end{itemize}

This paper is an extension of our previous paper~\cite{yuan2025local}. Building upon the foundation established in~\cite{yuan2025local}, this paper introduces the following key contributions: 
(i) We propose a novel resource-aware adaptation mechanism that fundamentally extends our framework's capability to handle arbitrary user-specified resource budgets. By incorporating resource constraints directly into the state space and employing randomized resource budgets during training, our approach enables robust generalization to diverse budget specifications unseen during training, addressing the practical challenge of dynamic resource requirements in real-world deployment scenarios.
(ii) We conduct extensive additional experiments analyzing different budget distribution strategies (uniform, normal, and hybrid) and their impact on training performance and generalization capability. Through comprehensive ablation studies, these experiments reveal how resource awareness in the state space and randomized budget training contribute to satisfying arbitrary post-training resource constraints, achieving near-zero constraint violations across diverse budget configurations.
(iii) We provide an in-depth exposition of our \DatasetName dataset, including detailed descriptions of the generation principles, comprehensive characterization of the multi-modal, multi-task, and multi-turn conversational data, and complete generation procedures with systematic quality control mechanisms.

\subsection{Organization}
The remainder of this paper is organized as follows. Sec. \ref{Sec. Related Work} reviews the related work. Sec. \ref{Sec. Proposed System} presents our proposed \Name system, and the design of RCRL is described in Sec.~\ref{Sec. RL}. Sec. \ref{Sec. Dataset} introduces our \DatasetName dataset and describes the data generation methodology. Sec. \ref{Sec. Experiment} details the experimental setup and results. Finally, Sec. \ref{Sec. Conclusion} concludes the paper and discusses future research directions.

\section{Related Work}
\label{Sec. Related Work}

\subsection{Efficient On-Device and Over-Network LLMs}
The rapid development of LLMs has profoundly influenced human-AI interaction while introducing significant challenges related to operational costs, energy consumption, and environmental impact. Researchers propose various solutions to enable deployment in resource-constrained devices, such as architectural optimization \cite{chu2023mobilevlm}, quantization \cite{lin2024awq}, on-device speculative decoding \cite{xu2024edgellm}, and adaptive deployment across heterogeneous devices \cite{zhang2025tensallo}. Furthermore, network-based approaches have emerged, such as \cite{shen2024large} proposed the client-edge co-inference method which partitions LLM into lightweight layers running on devices and parameter-heavy layers operating at the edge, collaborative inference on edge devices \cite{ye2025jupiter}, and edge inference optimization in wireless networks \cite{zhang2024beyond} to enhance inference efficiency.

However, existing frameworks primarily focus on either single-modal data sources or single-query accuracy \cite{zhang2024efficient}, neglecting the complex trade-off optimization of key resource indicators (e.g., response quality, latency, costs) \cite{dong2024creating} and failing to address the intricate relationships between multi-modal, multi-task, and multi-turn conversation context (i.e., RQ1 and RQ2). Furthermore, \cite{zhang2024efficient, he2024large} employ datasets with fixed evaluation metrics, which may not fully represent the complexity in many real-world LLM applications. Moreover, online training methods incur prohibitively expensive costs and lead to irreproducible resource consumption, particularly when considering human subjects or alternative approaches for evaluating LLM inference (i.e., RQ4). In contrast to previous work, we propose an offline training approach to preserve LLM inference results. While this inevitably introduces uncertainties in the LLM inference data, we address this challenge by further incorporating an uncertainty estimator for response scores.

\subsection{Inference Routing and Offloading Optimization}

More generally, there is a rich set of literature on strategies for optimizing device-edge-cloud inference location selection/offloading, based on metrics such as performance, cost, and latency \cite{singhal2024resource}. These works explore diverse methodological approaches that can be categorized into network optimization strategies, such as confidence-based device-server selection \cite{al2024regret} and energy-delay considerations for LLM training \cite{liu2024resource}, as well as RL frameworks, such as accuracy-delay-fairness device-server selection \cite{beytur2024optimization} and multi-device offloading with quality-latency-energy considerations \cite{chen2024tilesr}. Recent works have also explored LLM routing strategies, such as multi-LLM routing \cite{zhang2025router}, self-routing \cite{fang2025collaborative}, and agentic routing \cite{fang2026iterative}.

However, existing hierarchical inference approaches \cite{al2024regret, beytur2024optimization}, which orchestrate offloading between user devices and cloud servers, only consider model size differences. They neglect fundamental distinctions in their input modalities, for example, one scenario is device-side LLMs being limited to text-only inputs, while server-side models can process multimodal inputs (i.e., RQ1). Furthermore, recall that the unique challenge in our settings is multi-turn conversation interactions, yet these methods primarily focus on meeting specific performance standards or finding \textit{immediate} optimal solutions without considering \textit{cumulative} reward maximization throughout the decision-making process. Recent work on active inference for LLM inference offloading \cite{he2024large} disregarded multi-modal data and multi-task scenarios in multi-turn conversations, failing to align with practical LLM applications (i.e., RQ2). Similarly, LLM routing approaches \cite{zhang2025router} focus on single-query routing without considering multimodal inputs, multi-turn conversations, or resource constraints. Moreover, some offloading methods \cite{fang2025collaborative} preset fixed offloading ratios that cannot be adjusted post-training, which fundamentally conflicts with users' expectations for flexibly adapting resource constraints in real-world scenarios (i.e., RQ3). Differently from prior works, we implement these concepts in a multi-turn conversation LLM scenario, not only maximizing cumulative rewards and considering the modality-task associations to select optimal LLMs and multi-modal data sources, but also ensuring robust generalization to arbitrary user-specified resource budgets.

\begin{figure*}[t]
    \centering
    \includegraphics[width=1\linewidth]{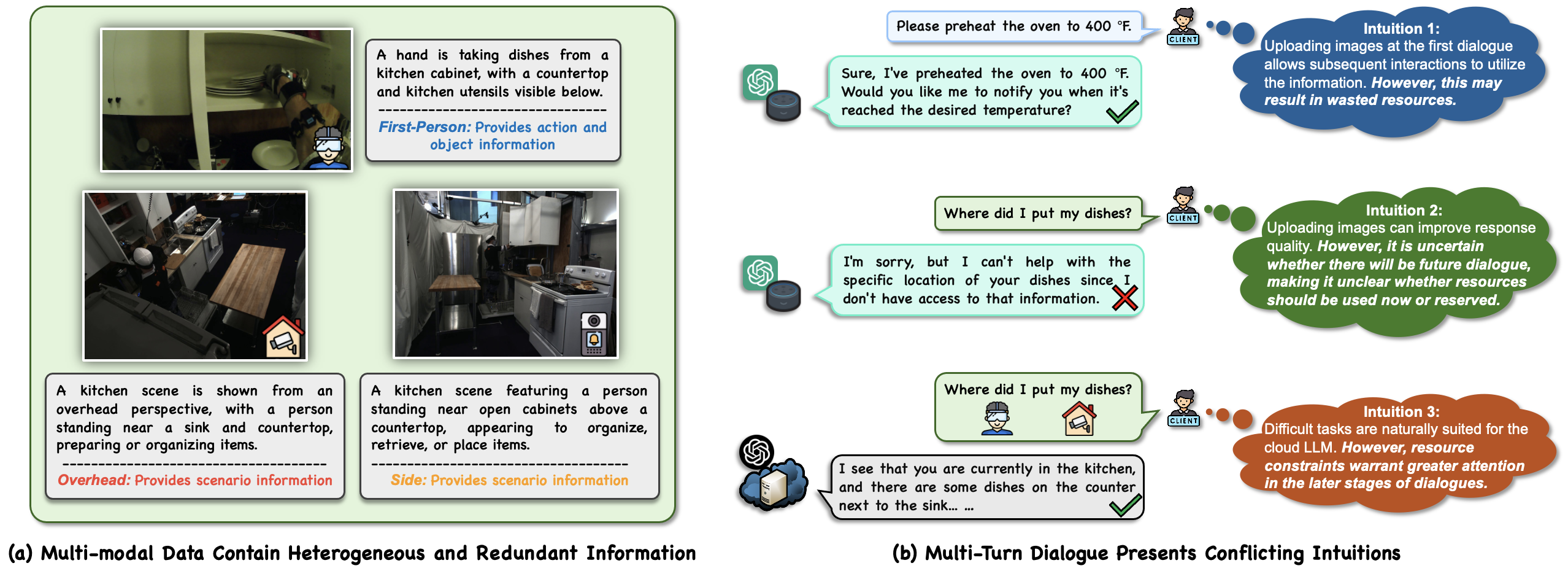}
    \caption{\textbf{Challenges Emerging in the \Name Scenario.} (a) Multi-modal data from different perspectives provide complementary information but also contain redundancy. (b) Multi-turn conversations present conflicting intuitions under resource constraints: early image uploading may waste resources or improve quality, but uncertainty about future turns makes optimal resource allocation unclear.}
    \label{Fig. Challenges}
\vspace{-4mm}
\end{figure*}

\section{Overview of the Proposed \Name System}
\label{Sec. Proposed System}

\subsection{Multi-Task Multi-Turn Scenario with Multi-Modal Data}
\label{Sec. Multi-Task Multi-Turn with Multi-Modal Data Sources}

We consider a setup in which a user is aiming to solve multiple types of inference tasks through a multi-turn conversation with LLMs, as shown in Fig. \ref{Fig. TMO}. Each conversation turn $t$ involves:
\begin{itemize}[leftmargin=5mm]
\item A text prompt $P_t$ which describes the user's intent, e.g., ``Recommend dishes that I can cook with my ingredients,'' and
\item A set of other modalities $\mathcal{M}_t \subseteq \mathcal{M}$, where $\mathcal{M}$ represents all modalities available in the current context (e.g., different image views, 3D points, or other data formats describing the current situation).
\end{itemize}
During the conversation with LLMs, a user may ask additional questions using prompts potentially related to the previous turns (e.g., ``Where did I put my dishes?'' or ``Please turn on all the lights in the kitchen''). Here, we assume each text prompt belongs to one of $N$ different task categories (e.g., assistant, recommendation, query, and message editing). Moreover, distinct tasks have varying levels of difficulty and require different types of multi-modal information to solve. For example, the prompt ``Where did I put my dishes?'' requires additional modalities (e.g., images) for the LLM to solve the task, while the prompt ``Please turn on all the lights in the kitchen'' may not require other sources of data, indicating that the task is simpler. These characteristics within a multi-task, multi-turn scenario with multi-modal data sources distinguish our setup from existing works \cite{zhang2024efficient, he2024large, yang2024perllm} that consider simpler settings (e.g., single-turn conversations with text only), motivating the need for a new solution.

\subsection{Device-Cloud LLM Inference}
\label{Sec. Device-Cloud LLM Inference}

Due to the inherent limitations of on-device hardware in terms of computational resources, memory, and energy consumption, deploying large-scale LLM directly on these devices is often impractical. On the other hand, relying exclusively on cloud-based LLMs can introduce latency and potential usage costs, especially when handling multi-modal data. We propose the device-cloud collaborative LLM inference (\Name) system, as shown in Fig. \ref{Fig. TMO}, which strategically distributes tasks between device and cloud platforms to exploit their respective strengths. It consists of two different LLMs: the on-device LLM and the cloud LLM. The \textbf{on-device LLM} is a lightweight LLM (e.g., Phi-3-mini \cite{abdin2024phi}) deployed on user devices to handle simple tasks efficiently. This model is optimized for high-speed text prompt processing, suitable for tasks such as message editing, setting timers, or adjusting air conditioning settings, where a sophisticated understanding of multi-modal data is not critical. On the other hand, the \textbf{cloud LLM} is a large-scale multi-modal LLM (e.g., GPT-4o \cite{OpenAI}) situated in the cloud capable of managing complex task and multi-modal data that exceed the processing capabilities of user devices. This LLM can integrate multi-modal data input, thus providing richer, more accurate responses where required. The advantage of the \Name approach is its flexibility and efficiency. In the \Name system, the user can flexibly choose to either offload inference to the cloud or process it on-device, depending on the task's difficulty and current resource requirements.
By offloading computationally intensive tasks to the cloud, we preserve on-device resources while still benefiting from the cloud's powerful computational capabilities. This configuration allows each part of the system to operate within its respective capacity, optimizing overall performance and user experience (e.g., latency and costs).

Formally, we describe the interaction with either the on-device or cloud LLM at each conversation turn $t$ as follows:
\begin{equation}
    R_t = \begin{cases}
    \text{LLM}_\text{Device}\left(P_t\right), \ \ \ \text{for device}, \\
    \text{LLM}_\text{Cloud}\left(P_t\right), \ \ \ \ \ \ \text{for cloud text-only}, \\
    \text{LLM}_\text{Cloud}\Big(P_t, \bigcup\limits_{m\in \mathcal{M}_t} \{D^{(m)}\}\Big), \\ \ \ \ \ \ \ \ \ \    \text{for cloud with modalities in  $\mathcal{M}_t$},
    \end{cases}
\label{Eq. LLM_1}
\end{equation}
where $R_t$, $P_t$, and $\mathcal{M}_t\subseteq \mathcal{M}$ represent the LLM response, text prompt, and the set of selected modalities uploaded to the cloud, respectively, at conversation turn $t$. $D^{(m)}$ denotes the data source of modality $m\in\mathcal{M}$, while $\text{LLM}_\text{Device}$  and $\text{LLM}_\text{Cloud}$ indicate the on-device LLM and the cloud LLM, respectively. LLMs can maintain contextual memory over extended multi-turn conversations, thereby generating responses for the current prompt based on the accumulated context of all previous prompts and responses. Hence, Eq. \eqref{Eq. LLM_1} can be rewritten to reflect the cumulative nature of the prompts. For example, in the case of the cloud LLM with multi-modal inputs,
\begin{equation}
R_t = \text{LLM}_\text{Cloud} \left(P_t, \bigcup_{i=0}^{t}\bigcup_{m \in \mathcal{M}_i}\{D^{(m)}\}, \bigcup_{i=0}^{t-1} \{P_i, R_i\}\right),
\label{Eq. LLM_2}
\end{equation}
where $t=1, 2, \dots, T$. In Eq. \eqref{Eq. LLM_2}, in addition to the inputs in Eq. \eqref{Eq. LLM_1}, we consider the previously selected modalities as well as prior prompts and responses from the series of earlier conversation turns $i=0, 1, \dots, t-1$.

\subsection{Challenges}
\label{Sec. Challenges}

\noindent\textbf{\Name Objective and Rationale.}
The primary objective of \Name is to strategically decide (i) whether to use the on-device LLM or cloud LLM and (ii) which multi-modal data sources to utilize, for each task within multi-turn conversations. This decision-making process aims to enhance response quality while efficiently managing latency and costs throughout extended interactions. The goal is to ensure high response quality in complex multi-turn conversational environments, particularly under the system constraints (e.g., latency and costs).

\noindent\textbf{Why \Name Decision-Making is Fundamentally Difficult?}
The difficulty of LLM and modality selection stems from the intricate interplay of multiple competing objectives. Fig.~\ref{Fig. Challenges} summarizes the key challenges of our problem. 
First, multi-modal data selection presents inherent challenges due to the heterogeneous nature of available data sources. As illustrated in Fig. \ref{Fig. Challenges}(a), different viewpoints (e.g., first-person, overhead, and side cameras) capture complementary information about the same scenario -- each perspective offers unique insights into actions and context. However, these multi-modal sources also contain substantial redundancy. For instance, while overhead and side views provide scenario information from different angles, they also capture overlapping visual content of the same objects and spatial layout. This creates a fundamental tension: while incorporating multiple modalities can enhance understanding, transmitting redundant information wastes communication resources and increases latency.
Second, selecting the appropriate LLM and modalities across consecutive turns introduces compounding complexity, as shown in Fig. \ref{Fig. Challenges}(b). For example, iPhone users encounter choices among multiple LLMs (e.g., the on-device Siri and the ChatGPT application), while also navigating various multi-modal data sources on mobile devices (e.g., front camera, rear camera, and screenshots). While individual queries may appear straightforward (e.g., using on-device LLM for simple message editing), three conflicting objectives emerge in multi-turn scenarios under resource constraints: (1) uploading images early enables richer context for subsequent turns but may waste resources if the conversation ends or future queries are simple; (2) deferring uploads preserves resources but forgoes early context benefits; (3) difficult tasks naturally warrant cloud LLM usage, yet cumulative resource consumption across turns complicates when to commit expensive operations. The decision-maker must simultaneously balance response quality, latency, and cost not just for individual turns, but cumulatively throughout extended conversations. This multi-objective optimization across multi-modal, multi-task, and multi-turn scenarios presents a formidable decision-making challenge.

\section{RL-Based Decision Engine in \Name}
\label{Sec. RL}

\begin{figure}[t]
    \centering
    \includegraphics[width=1\linewidth]{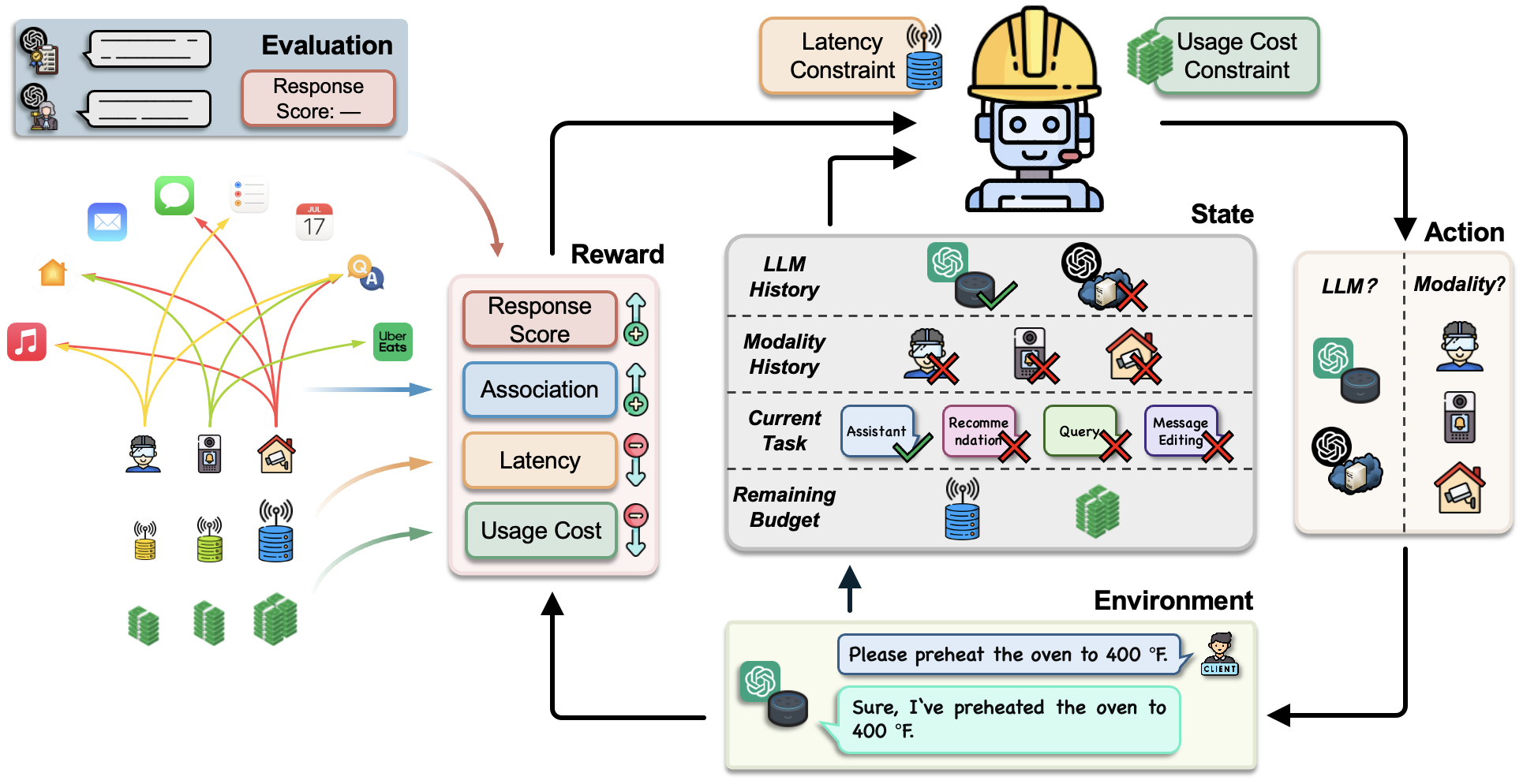}
    \caption{\textbf{Illustration of RCRL within the \Name System.} The reward signal relies on the reference and judge LLMs for the response score and the pre-trained CLIP model for the association, which together account for the dominant cost of training. These auxiliary invocations are not required at deployment, where the RL policy outputs actions from the state alone.}
    \label{Fig. RL}
\vspace{-4mm}
\end{figure}

We propose a resource-constrained RL (RCRL) approach to make informed decisions regarding the selection between on-device and cloud LLM services, as well as the determination of which data modalities to be uploaded, as shown in Fig. \ref{Fig. RL} and Alg. \ref{alg:system}. The details are described in the following.

\subsection{RL Formulation}
\label{Sec. Formulation}
We first formulate different components of the RL policy as follows. 

\noindent\textbf{\textit{State:}} 
The state captures the history of previous LLM selections, uploaded modalities, and task categories over a sliding window of the most recent $\tau$ turns. Specifically, for each turn $t$, the state $s_t$ records: (i) the selected LLM type in the previous turn $\ell_{t-1} \in \{\text{LLM}_\text{Device}, \text{LLM}_\text{Cloud}\}$, (ii) the set of modalities $\mathcal{M}^{\ell}_t$ available to the selected LLM (where $\mathcal{M}^{\ell}_t = \mathcal{M}_t$ if cloud LLM has been selected, and $\mathcal{M}^{\ell}_t = \emptyset$ for on-device LLM which does not support multi-modal inputs), and (iii) the task category $n_t \in \{1,...,N\}$. The state space $\mathcal{S}$ maintains a fixed-length history of $\tau$ consecutive turns, with each historical entry represented as a tuple $(\ell, \mathcal{M}^{\ell}, n)$. Formally, the state at turn $t$ is constructed as:
\begin{equation}
s_t = [(\ell_{t-\tau}, \mathcal{M}^{\ell}_{t-\tau+1}, n_{t-\tau+1}), \ldots, (\ell_{t-1}, \mathcal{M}^{\ell}_t, n_t)],
\label{Eq. State}
\end{equation}
where each tuple corresponds to one turn in the history window. For turns where $t < \tau$ (i.e., insufficient history is available), we pad the state with placeholder values (denoted as $-1$) to maintain a consistent dimensionality. Concretely, the state evolution can be illustrated as follows:
\begin{equation}
\begin{aligned}
s_0 &= [\underbrace{-1, \ldots, -1}_{(\tau-1) \text{ padding entries}}, (\ell_{-1}, \mathcal{M}^{\ell}_0, n_0)], \\
s_1 &= [\underbrace{-1, \ldots, -1}_{(\tau-2) \text{ padding entries}}, (\ell_{-1}, \mathcal{M}^{\ell}_0, n_0), (\ell_0, \mathcal{M}^{\ell}_1, n_1)], \\
&\vdots \\
s_t &= [(\ell_{t-\tau}, \mathcal{M}^{\ell}_{t-\tau+1}, n_{t-\tau+1}), \ldots, (\ell_{t-1}, \mathcal{M}^{\ell}_t, n_t)]  \\
& \qquad \qquad \qquad \qquad \qquad \qquad \qquad \qquad \text{for } t \geq \tau-1,
\end{aligned}
\label{Eq. State Evolution}
\end{equation}
where $\ell_{-1}$ denotes a placeholder value (i.e., $-1$) indicating no prior action has been taken. As the episode progresses, the padding is gradually replaced by actual historical observations until the full window of $\tau$ turns is populated.

\noindent\textbf{\textit{Action:}} 
The action taken for the next turn consists of the choice between utilizing on-device or cloud LLM services, as well as the selection of one or multiple data modalities for inference processing. Therefore, the action space $\mathcal{A}$ can be described as follows:
\begin{equation}
    \mathcal{A} = (\ell, \mathcal{M}^{\ell}),
\end{equation}
with the action $a_t = (\ell, \mathcal{M}^{\ell})_t$ for turn $t$ consisting of a choice of LLM and modalities to upload. For notational simplicity, we will refer to $\mathcal{M}^{\ell}$ as simply $\mathcal{M}$ when cloud LLM is selected with the corresponding modalities, while for on-device LLM, $\mathcal{M}^{\ell}$ is always empty (text-only input).

\noindent\textbf{\textit{Reward:}} In designing the reward function, we consider several key metrics to optimize the decision-making process regarding the choice of LLM service and modalities. Mathematically, the reward function $\mathcal{R}: \mathcal{S} \times \mathcal{A} \rightarrow \mathbb{R}$ that we propose is expressed as follows:
\begin{equation}
r_t = \begin{cases}
\begin{aligned}
&\alpha S_{a_t} + \beta_{\Lambda} \Lambda_{a_t} - \beta_{\psi} \psi_{a_t} - \beta_\phi \phi_{a_t}, \\
&\qquad\qquad\quad \text{if} \quad \sum_{i=\max(0, t-\tau)}^{t-1} C_j(a_i) \leq \xi_j, \; \forall j \in \mathcal{J}
\end{aligned} \\
\\
0, \qquad\qquad \text{otherwise}
\end{cases}
\label{eq:reward}
\end{equation}
where:
\begin{itemize}[leftmargin=5mm]
    \item The \textbf{response score} $S_{a_t}$ assesses how effectively the LLM response meets the requirements of the text prompt. Here, the response is generated by the LLM service selected in action $a_t$ using the chosen modalities. The response score is used solely for RL training purposes and is not required during the inference phase. A more detailed description of this metric and how we estimate it is provided in Sec.~\ref{Sec. Response Score}.
    \item The \textbf{association metric} $\Lambda_{a_t}$ quantifies the aggregated relevance of the multi-modal data to prompt $P_t$ for the modality set selected in action $a_t$. Formally, $\Lambda_{a_t} = \sum_{m \in \mathcal{M}_t} \Lambda(P_t, D^{(m)})$, where $\mathcal{M}_t$ is the set of modalities uploaded based on action $a_t$. Further details are provided in Sec. \ref{Sec. Association}.
    \item The \textbf{latency} $\psi_{a_t}$ represents the inference latency resulting from action $a_t$. This includes the computation time of the on-device LLM or the communication and processing time with the cloud LLM for the selected data modalities. Refer to Sec. \ref{Sec. Latency Model} for the formal definition.
    \item The \textbf{usage cost} $\phi_{a_t}$ reflects the cost incurred by action $a_t$, including the energy consumption of the on-device LLM or the service fee of the cloud LLM for the uploaded modalities, as described in Sec. \ref{Sec. Usage Cost Model}.
    \item The \textbf{resource constraints} $\xi_j$ ensure that the cumulative consumption $C_j$ of each resource type $j \in \mathcal{J}$ over a time window $\tau$ does not exceed the user-specified budgets $\xi_j$. If any constraint is violated, the reward is set to zero to penalize infeasible actions, as detailed in Sec. \ref{Sec. Resource Awareness}.
\end{itemize}
The response score $S_{a_t}$ and the association metric $\Lambda_{a_t}$ are positive indicators (higher is better, $\uparrow$). In contrast, latency $\psi_{a_t}$ and usage cost $\phi_{a_t}$ are negative indicators (lower is better, $\downarrow$). All metrics are normalized to the range $[0, 1]$ to ensure commensurability. Variable $\alpha$ controls the overall weight of the response score, while $\beta_\Lambda$, $\beta_\psi$, and $\beta_\phi$ are weighting factors for the remaining components satisfying $\beta_\Lambda + \beta_\psi + \beta_\phi = 1$.

\makeatletter
\algrenewcommand\algorithmiccomment[1]{\hfill$\triangleright$~#1\unskip}
\makeatother

\begin{algorithm*}[t]
\small
\caption{RCRL Training with \DatasetName in Our \Name System}
\label{alg:system}
\textbf{Input:} M4A1 Dataset ($\mathcal{D}$), initial policy parameters ($\pi_\theta$), reward weights ($\alpha$, $\beta_{\Lambda}$, $\beta_\psi$, $\beta_\phi$), discount factor ($\gamma$), time span ($\tau$), number of neighbors for response score estimation ($k$), learning rate ($\eta$), budget distributions ($\{\mathcal{B}_j\}_{j \in \mathcal{J}}$) \\
\textbf{Output:} Refined policy $\pi_\theta$ for LLM and modality selection

\begin{algorithmic}[1]
\For{each training iteration}
    \State Sample an episode from $\mathcal{D}$ including a sequence of states $\{s_t\}_{t=1}^{T}$ with all association metrics $\{\Lambda_a\}$. \Comment{Alg.~\ref{alg:dataset}}
    \State Sample initial resource budgets: $\xi_j \sim \mathcal{B}_j$ for all $j \in \mathcal{J}$. \Comment{Eq.~\eqref{eq:budget_sampling}}
    \For{each time step $t$ in the episode}
        \State Construct resource-aware state $s_t^\text{RA}$ by concatenating remaining budget $\boldsymbol{\xi}_t$. \Comment{Eq.~\eqref{eq:resource_awareness}}
        \State Predict action $a_t \sim \pi_\theta(s_t^\text{RA})$ via the current policy.
        \State Estimate uncertain response score $S_{a_t}$ for state-action pair $(s_t^\text{RA}, a_t)$ via nearest neighbors. \Comment{Eq.~\eqref{eq:ResponseScore}}
        \State Retrieve the association metric $\Lambda_{a_t}$ for action $a_t$ from the sampled episode data.
        \State Compute the device-side latency $\psi_{a}$ or record the real interaction time with the cloud LLM provider. \Comment{Eq.~\eqref{eq:LocalLatency}}
        \State Compute the usage cost $\phi_{a_t}$ for either the on-device or cloud LLM. \Comment{Eqs.~\eqref{eq:LocalUsageCost},~\eqref{eq:CloudUsageCost}}
        \State Compute reward $r_t$ based on $S_{a_t}$, $\Lambda_{a_t}$, $\psi_{a_t}$, and $\phi_{a_t}$. \Comment{Eq.~\eqref{eq:reward}}
        \State Update remaining budget $\boldsymbol{\xi}_{t+1}$ based on resource consumption from action $a_t$.
    \EndFor
    \State Calculate the RL loss $f(\theta)$ using the episode trajectory $\{(s_t^\text{RA}, a_t, r_t)\}_{t=1}^{T}$. \Comment{Eq.~\eqref{eq:A2Closs}}
    \State Update the policy parameters: $\theta \leftarrow \theta - \eta \nabla_\theta f(\theta)$.
\EndFor
\end{algorithmic}
\end{algorithm*}

\subsection{MDP and Optimization}
\label{Sec. MDP Optimization}
Based on our state and action definitions, we model this problem as a Markov decision process (MDP), where the learning objective is to maximize the cumulative reward. At each conversation turn $t$, RL algorithms observe the state $s_t$, select an action $a_t$, obtain a transition probability $P(s_{t+1} | s_t, a_t)$, receive a reward $r_t$, and transition to the next state $s_{t+1} \sim P(\cdot | s_t, a_t)$. Given the current state $s_t$, the action $a_t$ is selected according to a specific policy $\pi$ as $a\sim\pi(\cdot|s)$, where $\pi(a|s)$ represents the probability of selecting action $a$ in state $s$. The value function $V^\pi(s)$ predicts the expected discounted rewards of states following the policy $\pi$:
\begin{equation}
\small
V^\pi(s) = \mathbb{E}_\pi \left[ \sum_{t=0}^T \gamma^t r_t \bigm| s_0 = s \right]
\end{equation}
with discount $\gamma > 0$. The optimization of policy $\pi_{\theta}$ with respect to its parameterization $\theta$ aims to maximize the expected discounted rewards in LLM inference offloading and modality selection:
\begin{equation}
\small
\max_\theta J(\theta) \triangleq \mathbb{E}_{s,a \sim \pi_{\theta}} \left[\sum_{t=0}^T \gamma^t r_t\right].
\end{equation}
In deep RL optimization, various strategies can be employed to adjust the parameters $\theta$ of the policy $\pi_\theta(a|s)$ with the goal of maximizing $J(\theta)$. To achieve this, a loss function $f(\theta)$ is defined such that minimizing $f(\theta)$ leads to maximizing $J(\theta)$. Taking Advantage Actor Critic (A2C) \cite{mnih2016asynchronous} as an example, the loss function $f_{\text{A2C}}(\theta)$ includes both the policy loss and the value loss, along with an entropy term to encourage exploration:
\begin{equation}
\begin{aligned}
f_{\text{A2C}}(\theta) &= \mathbb{E}_{s,a \sim \pi_\theta} [ -\log \pi_\theta(a|s) A(s, a) \\
&+ \frac{1}{2} (V^{\pi_\theta}(s) - r)^2 - \zeta H(\pi_\theta(\cdot|s)) ],
\end{aligned}\label{eq:A2Closs}
\end{equation}
where $A(s, a) = r + \gamma V^{\pi_\theta}(s') - V^{\pi_\theta}(s)$ for $s' \neq s$ is an estimate of the advantage function, providing a measure of the relative value of taking action $a$ in state $s$. The coefficient $\zeta$ controls the weight of the entropy bonus $H(\pi_\theta(\cdot|s))$. In Sec.~\ref{Sec. Experiment Setup}, we also consider Proximal Policy Optimization (PPO) \cite{schulman2017proximal} and Deep Q Network (DQN) \cite{mnih2015human} versions of Eq.~\eqref{eq:A2Closs}.

\subsection{Uncertain Response Score Estimation}
\label{Sec. Response Score}

Due to the vast number of parameters in LLMs, offline training approaches have become essential to mitigate substantial inference costs and enable future reusability \cite{ouyang2022training}. Similarly, \Name's RCRL procedure requires a specialized dataset for learning and exploration of LLM performance in multi-modal, multi-task, and multi-turn conversational settings. Moreover, given the nature of multi-task settings, particularly in generative tasks, there are often no explicit evaluation metrics, such as accuracy, task success rate, or error rate. This creates a significant challenge in assessing the quality of LLM responses. Furthermore, the exploratory nature of RL, which frequently encounters new state-action pairs, presents significant evaluation challenges.

To address these issues, we propose leveraging the notion of \textit{LLM-as-Judge} to evaluate response quality. However, LLM-based judges exhibit inherent scoring instability, where identical responses may receive different scores across evaluations. To handle this uncertainty, we employ a nearest neighbor strategy to estimate and stabilize response scores for outputs from four LLMs: the on-device LLM, the cloud LLM, a reference LLM, and a judge LLM.

\subsubsection{Response Quality Evaluation}
The response score $S_{a_t}$ measures the response quality generated by the selected LLM at each conversation turn. In multi-task scenarios, evaluating the response quality requires assessing the actual utility of responses in aiding human users, particularly in query tasks. We employ the quantitative evaluation scheme outlined by \cite{liu2024visual}, illustrated on the right side of Fig.~\ref{Fig. M4A1}. This involves using two additional \textit{text-only} LLMs specifically to assess the quality of candidate LLM services:
\begin{itemize}[leftmargin=5mm]
\item The \textit{reference LLM} processes a prompt and a reference response to produce a polished reference response, serving as a benchmark for evaluation.
\item The \textit{judge LLM} then evaluates both the candidate's response and the reference's response, scoring them based on helpfulness, relevance, accuracy, and detail.
\end{itemize}
The response score $S_{a_t}$ for turn $t$ is then calculated as the ratio of the candidate's LLM score to the reference score.

\subsubsection{Handling Uncertainty in Response Score Estimation}
Integrating RCRL directly into real-time LLM inference is impractical, especially in our scenario, where four LLMs operate in tandem: on-device and cloud LLMs generate responses, while reference and judge LLMs evaluate their quality. Consequently, we archive the LLM inference processes into a dataset for RCRL training. However, this approach introduces two main forms of uncertainty:
\begin{itemize}[leftmargin=5mm]
\item[(i)] \textit{Non-deterministic evaluation (NDE)}: LLM-based judges exhibit inherent scoring instability, where identical state-action pairs (i.e., the same prompt and response) may receive different quality scores across multiple evaluations. This non-determinism stems from the stochastic nature of LLM inference, including sampling-based generation and variations in the judge LLM's internal reasoning processes.
\item[(ii)] \textit{Out-of-distribution (OOD) state-action pairs}: The dataset may contain state-action combinations that were never observed during data collection. For instance, certain combinations of LLM selections, input modalities, and task types may be underrepresented or entirely absent in the training data, making it challenging to accurately estimate their response quality.
\end{itemize}
In contrast to \cite{ran2023policy}, which aims to avoid selecting OOD actions, our goal is to more accurately estimate the response quality under OOD actions rather than constrain them. This adds flexibility for datasets where the states and actions are randomly generated (e.g., the current task, selected LLM, and selected modality), making it inevitable that some state-action pairs will not appear in the dataset (such as for our \DatasetName dataset described in Sec.~\ref{Sec. Dataset}).

To address these uncertainties, we estimate the response score $S'_{a}$ for a pending state-action pair $(s', a')$ by leveraging its nearest neighbors in the dataset. Specifically, let $\{(s^i, a^i)\}_{i=1}^k$ be the $k$ state-action pairs nearest to $(s', a')$ in Euclidean distance, and let $\{S_i\}_{i=1}^k$ denote their known response scores. We estimate $S'_{a}$ as a weighted average over its neighbors:
\begin{equation}
\small
S'_{a} = \frac{\sum_{i=1}^k {S_i/d_i}}{\sum_{i=1}^k 1/d_i}, \quad \mathrm{where} \quad d_i = \left\|[s' \; a'] - [s^i \; a^i]\right\|
\label{eq:ResponseScore}
\end{equation}
represents the distance between the $k$ nearest state-action pairs and the pending state-action pair $(s', a')$, with $[x \; y]$ denoting vector concatenation. In this way, we achieve a more robust offline estimation of response quality, effectively handling both NDE and OOD uncertainties within RCRL for LLM inference.

\subsection{Task-Modality Association}
\label{Sec. Association}
\Name is designed to operate in multi-modal, multi-task scenarios. In such settings, quantifying the relevance between users' text prompts and multi-modal data sources enables selection of the most pertinent multi-modal data to offload to the cloud LLM. Consequently, we employ feature extractors to transform both tasks and modalities into feature vectors of identical dimensionality for similarity computation. Given this premise, we leverage the pre-trained CLIP model \cite{radford2021learning}, a transformer-based architecture with text and image encoders, to extract features from prompts and multi-modal data sources. The pre-trained CLIP has been extensively trained on a vast array of image-text pair data, effectively linking visual concepts with text. We employ a normalized cross-modality similarity metric to quantify the association between a text prompt $P$ and a modality set $\mathcal{M}$, which is defined as:
\begin{equation}
\Lambda = \sum_{m \in \mathcal{M}} \Lambda_m \left(P, D^{(m)}\right) = \sum_{m \in \mathcal{M}} \frac{\left\langle \omega_T(P), \omega_I(D^{(m)}) \right\rangle}{\left\|\omega_T(P)\right\| \cdot \left\|\omega_I(D^{(m)})\right\|},
\label{eq:Association}
\end{equation}
where $\omega_T: P \rightarrow \mathbb{R}^d$ and $\omega_I: D^{(m)} \rightarrow \mathbb{R}^d$ represent the text and image encoder functions, respectively, mapping inputs to a shared $d$-dimensional embedding space. Here, $\Lambda_m(P, D^{(m)})$ denotes the association metric between prompt $P$ and individual modality $m$, with normalization through $L_2$ norms producing similarity scores bounded in $[-1, 1]$. The total association metric $\Lambda$ is obtained by summing over all modalities in $\mathcal{M}$.

\subsection{Latency Model}
\label{Sec. Latency Model}

\subsubsection{On-Device Computation Latency $\psi_\text{Device}$}
The computational latency of the on-device LLM, or say inference time, is influenced not only by the number of LLM parameters, but also significantly by the length of the prompt, denoted $|P|$, and of the response, $|R|$. Following \cite{canziani2016analysis}, we assume that the computational latency is proportional to the number of parameters, though other latency models can be easily applied. If the on-device LLM is comprised of $|\text{LLM}_\text{Device}|$ parameters and requires $2|\text{LLM}_\text{Device}|$ floating point operations (FLOPs) for forward propagation, the computational demand for each token in a reference processing length $|P|_\text{ref}$ is calculated as $\frac{2|\text{LLM}_\text{Device}|}{|P|_\text{ref}}$ FLOPs. The latency calculation considers the theoretical peak performance of the user device's GPU, denoted as $\Theta_\text{peak}$, measured in TeraFLOPS (trillion floating-point operations per second). The computational latency can be computed as:
\begin{equation}
    \psi_\text{Device} = \frac{2|\text{LLM}_\text{Device}| (|P|+|R|)}{|P|_\text{ref} \times \Theta_\text{peak}}.
    \label{eq:LocalLatency}
\end{equation}

\subsubsection{Cloud Interaction Latency $\psi_\text{Cloud}$} 
The interaction time with the cloud LLM encompasses the entire process from sending the prompt to receiving the response. This includes uploading the prompt and modalities, inference using the cloud LLM, and downloading the response. We directly record the interaction time with the cloud LLM for different sets of uploaded modalities $\mathcal{M}$ during the creation of our \DatasetName dataset.

\subsection{Usage Cost Model} 
\label{Sec. Usage Cost Model}

During the inference process, LLMs incur computational costs. From the user's perspective, for the on-device LLM, this manifests as energy consumption, while for the cloud LLM, it translates into the service fee paid to the provider.

\subsubsection{On-Device Usage Cost $\phi_\text{Device}$}
The energy consumption of the on-device LLM is influenced by inference time and the power consumption of the computing hardware, which are in turn affected by the user device's computational capacity, the sizes of the prompt and response, and the size of the LLM, all encapsulated within the inference time $\psi_\text{Device}$. Let the maximum power consumption for a given user device hardware be $W_\text{max}$ Watts. By utilizing a normalized energy cost factor $\kappa$ (capturing e.g., local electricity rates), after appropriate unit conversion, the on-device LLM usage cost (in USD) is defined as:
\begin{equation}
    \phi_\text{Device} = \psi_\text{Device} \times W_\text{max} \times \kappa.
    \label{eq:LocalUsageCost}
\end{equation}

\subsubsection{Cloud LLM Usage Function $\phi_\text{Cloud}$}
For cloud-based LLMs, in addition to the inference costs of prompts and responses, the inference costs of data modalities must also be considered. These costs are typically available on the service provider's website. Taking GPT-4o as an example, the prompt cost rate is $\varphi_P = \$ 0.005 / 1K$ tokens, the response cost rate is $\varphi_R = \$ 0.015 / 1K$ tokens, and the data modality costs are proportional to their size and resolution at a rate $\varphi_m$ for modality $m \in \mathcal{M}$. Therefore, the usage cost of the cloud LLM can be defined as follows:
\begin{equation}
    \phi_\text{Cloud} = \varphi_P \cdot |P| + \varphi_R \cdot |R| + \sum_{m \in \mathcal{M}} \varphi_m \cdot \big|D^{(m)}\big|.
    \label{eq:CloudUsageCost}
\end{equation}

\subsection{Resource-Awareness and Constraint Generalization}
\label{Sec. Resource Awareness}
An essential objective of our approach is to ensure that users can arbitrarily adjust resource budget settings to match their preferences during real-world deployment, without requiring model retraining. However, achieving this goal presents significant challenges. Simply penalizing constraint violations in the reward function is insufficient, as it fails to provide the RCRL with adequate information about the resource costs incurred by actions and their relationship to the available resource budget. Moreover, training with fixed resource constraints during the learning process leads to overfitting, resulting in poor generalization to arbitrary user-specified budgets. To address these fundamental issues, we propose a dual-pronged strategy that simultaneously enhances the RCRL's awareness of resource consumption and its generalization capabilities across diverse constraint configurations.

First, we augment the state space to explicitly incorporate remaining resources as an integral component of the state representation, thereby enabling the RCRL to reason about resource availability when making decisions. Formally, we extend the original state space $\mathcal{S}$ defined in Eq. \eqref{Eq. State} to construct a resource-aware state space $\mathcal{S}^\text{RA}$. For each conversation turn $t$, the resource-aware state $s_t^\text{RA} \in \mathcal{S}^\text{RA}$ is constructed by concatenating the original state $s_t \in \mathcal{S}$ with a remaining budget vector $\boldsymbol{\xi}_t \in \mathbb{R}^{|\mathcal{J}|}$:
\begin{equation}
\begin{aligned}
s_t^\text{RA} = [s_t \; \boldsymbol{\xi}_t],& \\
\text{where } &\boldsymbol{\xi}_t = \left[\xi_j - \sum_{i=\max(0, t-\tau)}^{t-1} C_j(a_i)\right]_{j \in \mathcal{J}},
\end{aligned}
\label{eq:resource_awareness}
\end{equation}
where $[x \; y]$ denotes vector concatenation. Each component of $\boldsymbol{\xi}_t$ represents the remaining budget for the corresponding resource type $j$, accounting for resource consumption over the most recent $\tau$ conversation turns. Here, $\xi_j$ is the initial budget for resource type $j$, and $C_j(a_i)$ is the cost of action $a_i$ for that resource type. This remaining budget dynamically updates at each turn as actions consume resources, providing the RL policy with real-time awareness of resource availability within the sliding window.

Second, to prevent overfitting to specific budgets and cultivate robust generalization capabilities, we employ a curriculum of randomized resource budgets during the RCRL training process. In each episode, we randomly sample initial resource budgets from predefined distributions, exposing the RL policy to a diverse spectrum of constraint scenarios. This approach addresses a critical challenge in RCRL: policies trained with fixed budgets often fail to generalize when deployed under different resource availability conditions. By training across a distribution of budgets, the RL policy learns resource-aware decision-making strategies rather than budget-specific heuristics. Specifically, for each resource type $j \in \mathcal{J}$, the initial budget $\xi_j$ in each episode is sampled as:
\begin{equation}
\xi_j \sim \mathcal{B}_j,
\label{eq:budget_sampling}
\end{equation}
where $\mathcal{B}_j$ denotes a budget distribution for resource type $j$, such as uniform or normal distributions.

\subsection{Summary of Reinforcement Learning in \Name} 
Recall that in the RCRL framework of \Name, we employ a resource-aware state space $\mathcal{S}^\text{RA}$ to replace the original state space $\mathcal{S}$, thereby enabling the RL policy to explicitly reason about resource availability as defined in Eq.~\eqref{eq:resource_awareness}. The action space $\mathcal{A}$ remains unchanged to enable the RL policy to select both the appropriate LLM (on-device vs. cloud) and the relevant modality(ies) for upload. Revisiting our initially defined reward function in Eq.~\eqref{eq:reward}, we substitute the direct response score $S_{a_t}$ with the estimated response score $S_{a_t}'$ defined in Eq.~\eqref{eq:ResponseScore} to achieve a more precise representation. The association metric is quantified through Eq.~\eqref{eq:Association}. Both the latency formulations ($\psi_\text{Device}$ and $\psi_\text{Cloud}$) and usage cost formulations ($\phi_\text{Device}$ and $\phi_\text{Cloud}$) are incorporated as $\psi_{a_t}$ and $\phi_{a_t}$ in the reward function, respectively. 
During training, the RL learner collects on-policy rollouts in this environment, where action-dependent state components evolve deterministically via Eq.~\eqref{eq:resource_awareness}, the task category $n_{t+1}$ is sampled from the empirical distribution of \DatasetName, and the reward at each step is computed via Eq.~\eqref{eq:reward} using the estimated response score in Eq.~\eqref{eq:ResponseScore}, the analytical latency and usage cost in Eqs.~\eqref{eq:LocalLatency}-\eqref{eq:CloudUsageCost}, and the precomputed association in Eq.~\eqref{eq:Association}. See Alg.~\ref{alg:system} for the full procedure.

\begin{algorithm}[t]
\small
\caption{\DatasetName Dataset Generation}
\label{alg:dataset}

\textbf{Input:} Image dataset, instruction dataset, time span $(\tau)$

\textbf{Output:} \DatasetName dataset

\begin{algorithmic}[1]
\While{collecting data}
    \State Initialize an episode by randomly selecting a moment in the scenario and retrieving its multimodal images.
    \State Sample $T \in \{2, \dots, \tau\}$ at random
    \For{$t = 1$ to $T$}
        \State Sample a user prompt $P$ from the instruction dataset.
        \State Select either the on-device or the cloud LLM at random.
        \If{cloud LLM is selected}
            \State Sample modalities $\mathcal{M}_t \subseteq \mathcal{M}$ to upload.
        \EndIf
        \State Generate a response from the conversation context, prompt, and the data $D^{(m)}, m \in \mathcal{M}_t$. \Comment{Eqs.~\eqref{Eq. LLM_1},~\eqref{Eq. LLM_2}}
        \State Obtain a reference response via the reference LLM.
        \State Score the response with the judge LLM against the reference.
    \EndFor
    \For{each $P$ and $D^{(m)}$ in the episode}
        \State Compute association $\Lambda_m(P, D^{(m)})$ via pre-trained CLIP.
        \Statex \Comment{Eq.~\eqref{eq:Association}}
    \EndFor
\EndWhile

\end{algorithmic}
\end{algorithm}

\setlength{\fboxrule}{1pt}
\setlength{\fboxsep}{1.5pt}

\definecolor{LocalLLMbox}{HTML}{4CEEC7}
\definecolor{LocalLLM}{HTML}{D1FCF1}
\definecolor{CloudLLMbox}{HTML}{000000}
\definecolor{CloudLLM}{HTML}{E9E9E9}
\definecolor{Scorebox}{HTML}{CB6B5C}
\definecolor{Score}{HTML}{F3DBD7}

\definecolor{Assistantbox}{HTML}{2D5F8A}
\definecolor{Assistant}{HTML}{DFEDFA}

\definecolor{Recommendationbox}{HTML}{9E4472}
\definecolor{Recommendation}{HTML}{F6D7ED}

\definecolor{Querybox}{HTML}{356920}
\definecolor{Query}{HTML}{E0F4DA}

\definecolor{Messagebox}{HTML}{442B84}
\definecolor{Message}{HTML}{E5E2F8}

\section{\DatasetName Dataset}
\label{Sec. Dataset}

\subsection{Principles of \DatasetName Dataset Generation}

Current offline RL datasets can be broadly categorized into three types: random datasets, expert datasets, and mixed-quality datasets, such as D4RL \cite{fu2020d4rl}, RL Unplugged \cite{gulcehre2020rl}, and RoboMimic \cite{mandlekar2022matters}. These dataset types are designed to train RL policies through different learning paradigms. For instance, an expert dataset generated by human specialists contains predominantly optimal actions, making it well-suited for imitation learning where the RL policy simply needs to replicate expert behavior to accomplish the task successfully. However, as discussed in Sec. \ref{Sec. Challenges}, even humans struggle with decision-making in the \Name system. Recruiting human experts to generate the \DatasetName dataset presents a formidable challenge. While we may possess intuitive heuristics, such as using on-device LLMs for simple tasks and cloud LLMs for complex ones, we inevitably face difficulties in multi-modal and multi-turn conversation scenarios. The decision of whether to submit modal data earlier to provide richer context or to defer submission until necessary involves inherent trade-offs that lack clear expert consensus.
Consequently, we adopt a fundamentally different approach. We design the \DatasetName dataset with completely randomized actions, where both LLM selection and modality submission are randomly determined. This randomization ensures uniform coverage of the (LLM, modality, task) grid, so that the $k$-NN response-score estimator in Eq.~\eqref{eq:ResponseScore} is well-defined and low-variance at every action the policy may select. In contrast, an expert dataset would concentrate samples on a narrow region of the grid and leave the estimator unreliable for actions outside that region.

\subsection{Overview of Proposed \DatasetName}

\begin{figure}[t]
    \centering\includegraphics[width=1\linewidth]{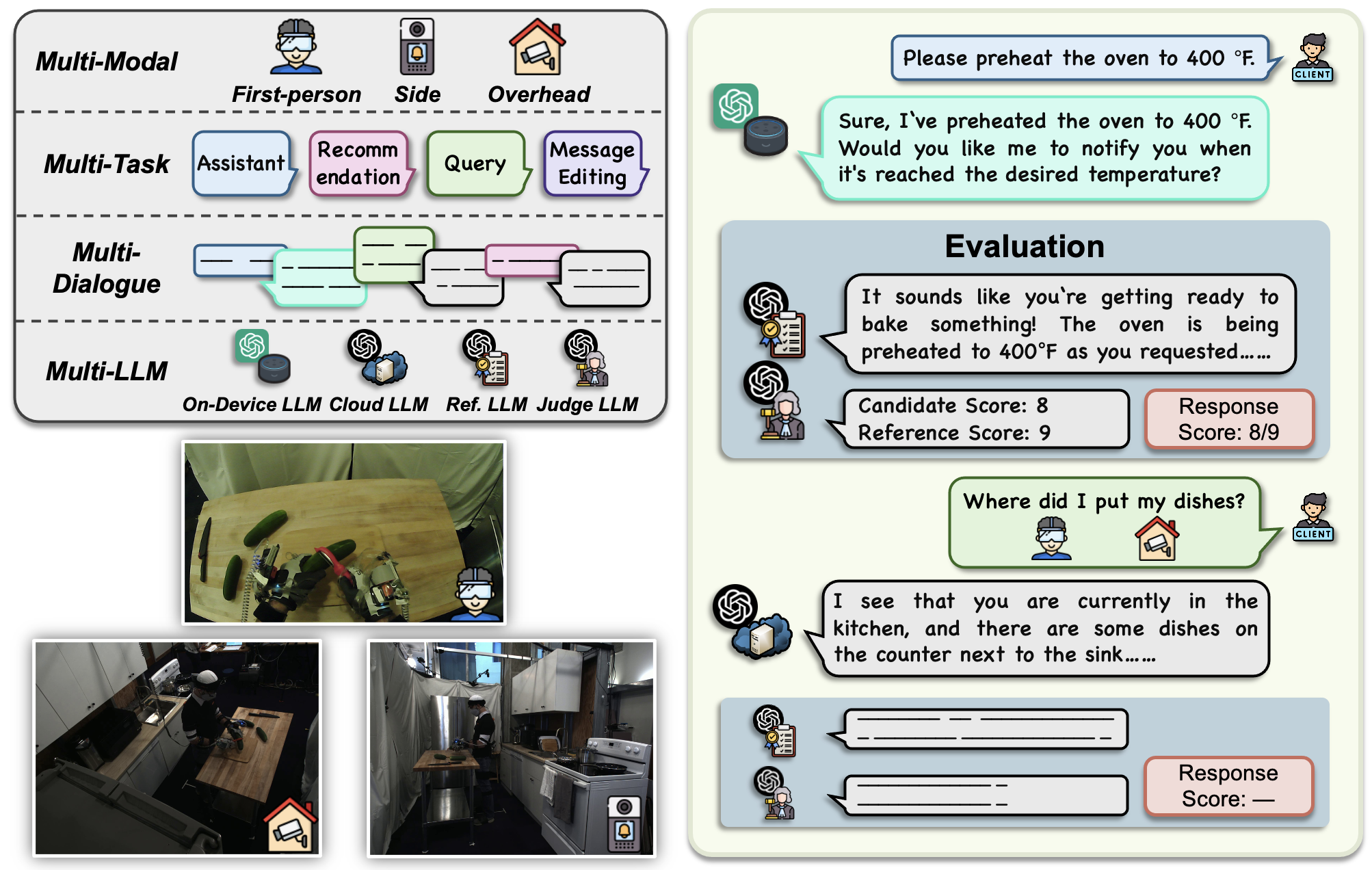}
    \caption{\textbf{Illustration of \DatasetName Dataset.} Each episode combines three image views (first-person, side, overhead), one of four tasks (Assistant, Recommendation, Query, Message Editing), multi-turn conversations, and four LLM roles (on-device, cloud, reference, judge), with the reference and judge LLMs scoring the candidate response.}
    \label{Fig. M4A1}
\vspace{-4mm}
\end{figure}

To evaluate performance in practical settings, we develop \DatasetName, the first multi-modal, multi-task, multi-turn, and multi-LLM dataset. The data collection method is detailed in Alg. \ref{alg:dataset}, and an overview of \DatasetName is depicted in Fig.~\ref{Fig. M4A1}. \DatasetName consists of 21,014 conversation turns, each meticulously compiled by randomly and sequentially choosing tasks, prompts, modalities, and candidate LLMs to ensure comprehensive coverage across different configurations. The key characteristics of \DatasetName can be summarized as follows:
\begin{itemize}
    \item \textbf{Three Multi-Modal Data Sources} consist of images from different views in a unified scenario, including first-person, side, and overhead views.
    \item \textbf{Four Tasks} include Assistant, Recommendation, Query, and Message Editing, represent varying levels of task difficulty, relevance to modalities, and the types of actions that the LLM will undertake.
    \item \textbf{Two to Five Conversation Turns} include pairs of text prompts and responses within a continuous context.
    \item \textbf{Four LLMs} contain a on-device LLM (Phi-3-mini \cite{abdin2024phi}), a cloud LLM (GPT-4o), a reference LLM (GPT-4o), and a judge LLM (GPT-4o \cite{OpenAI}).
    \item \textbf{Four Metrics} include response score (Sec. \ref{Sec. Response Score}), association (Sec. \ref{Sec. Association}), latency (Sec. \ref{Sec. Latency Model}), and usage cost (Sec. \ref{Sec. Usage Cost Model}).
\end{itemize}

\subsection{\DatasetName Generation Details}

To construct the \DatasetName dataset, we integrate two predetermined datasets that provide complementary modalities, images and text. First, we use the ActionSense dataset \cite{delpreto2022actionsense} to provide images, offering various camera views of human along with labels for each kitchen activity within the scenario, as illustrated in the bottom left of Fig. \ref{Fig. M4A1}. ActionSense is a large dataset of human kitchen activities that includes multi-modal wearable sensor data and environmental camera data, capturing 20 different kitchen activities such as peeling cucumbers, slicing potatoes, and cleaning plates with a sponge. These annotated kitchen activity labels function as responses to queries such as ``What activity am I doing right now?'' Given that current LLMs rarely process time-series sensor data (such as data from gyroscopes and accelerometers), we consider three types of image modalities: first-person images from a wearable sensor, and side and overhead views from environmental cameras. Second, the instruction dataset is used to provide structured text data, enriching the dataset with linguistic elements crucial for LLM inference and evaluation. This dataset provides text for four different tasks with prompts. Each sample in \DatasetName is constructed based on a set of three multi-modal data from ActionSense, paired with multiple randomly selected prompt and reference response pairs from the instruction dataset.

\section{Experiments}
\label{Sec. Experiment}

\subsection{Experiment Setup}
\label{Sec. Experiment Setup}

\noindent\textbf{Training Setup.} The \DatasetName dataset is split into a training set and a test set in an 8:2 ratio. The training set is used to train RLs while the test set is used to evaluate the performance at inference. When implementing our approach, we consider several popular discrete RL algorithms, including PPO \cite{schulman2017proximal}, DQN \cite{mnih2015human}, and A2C \cite{mnih2016asynchronous}. We consider these methods with and without resource constraints, and denote RC-PPO, RC-DQN, and RC-A2C as the methods with the resource constraints. The policies are parameterized by multilayer perceptrons (MLP), with a total of 30,000 time steps set for training. For each experiment, we repeat the process three times and report the standard deviation as the error bar. We implement the \Name system and the baselines in PyTorch and conducted the experiments on an NVIDIA A100 GPU with 40 GB of memory.

\newcommand{\com}[1]{\tiny$\pm$#1}
\newcolumntype{a}{>{\columncolor{gray!15}}c}

\begin{table*}[t]
\caption{Main Result - \Name with RC-A2C optimally balances on-device/cloud LLMs and adaptively selects modalities, outperforming in response quality, latency, and costs without constraint violations. The highlighted row indicates the highest overall reward without constraint violations.}
\vspace{-2mm}
\label{Table Main Result}
\centering 
\resizebox{\linewidth}{!}{
\begin{tabular}{l|ccc|a|ccccc|cc}
\toprule
\multirow{2}{*}{\textbf{Method}} & \textbf{Response} & \textbf{Latency} & \textbf{Usage Cost} & \textbf{Overall} & \multirow{2}{*}{\textbf{Device}} & \multicolumn{4}{c|}{\textbf{Cloud - Num. Selected Modalities}} & \multicolumn{2}{c}{\textbf{Constraint Violation} ($\downarrow$)} \\
& \textbf{Score} ($\uparrow$) & (s) ($\downarrow$) & (1e-3 USD) ($\downarrow$) & \textbf{Reward} ($\uparrow$) & & 0 (text-only) & 1 & 2 & 3 & Latency & Usage Cost \\
\midrule

On-Device & 0.74 \com{0.04} & 0.04 \com{0.00} & 0.00 \com{0.00} & 0.74 \com{0.04} & 4203 & 0 & 0 & 0 & 0 & \cellcolor{Yes} 0.00 \com{0.00} & \cellcolor{Yes} 0.00 \com{0.00} \\
Cloud & 1.04 \com{0.01} & 20.24 \com{0.18} & 25.14 \com{0.32} & 0.75 \com{0.01} & 0 & 515 & 1583 & 1540 & 544 & \cellcolor{No} 1.11 \com{0.05} & \cellcolor{No} 2.69 \com{0.34} \\
Random & 0.86 \com{0.02} & 10.00 \com{0.25} & 12.32 \com{0.23} & 0.71 \com{0.01} & 2097 & 261 & 775 & 771 & 255 & \cellcolor{No} 0.95 \com{0.13} & \cellcolor{No} 0.89 \com{0.83} \\
\bottomrule

\multicolumn{11}{c}{\textbf{\centering \rule{0pt}{10pt} LLM-as-Router \cite{ding2024hybrid, fang2025collaborative} for Inference Offloading (Ignored LLM Router's Latency and Usage Cost)}} \\ 
\toprule
Phi-3-mini & 0.81 \com{0.04} & 5.25 \com{0.31} & 7.54 \com{0.44} & 0.74 \com{0.04} & 3089 & 0 & 613 & 207 & 295 & \cellcolor{Yes} 0.00 \com{0.00} & \cellcolor{Yes} 0.00 \com{0.00} \\
Phi-3.5-mini & 0.83 \com{0.04} & 5.63 \com{0.38} & 8.39 \com{0.57} & 0.76 \com{0.04} & 3118 & 0 & 42 & 1044 & 0 & \cellcolor{Yes} 0.00 \com{0.00} & \cellcolor{Yes} 0.00 \com{0.00} \\
LLaMA-3.2-3B & 1.05 \com{0.02} & 22.04 \com{0.27} & 35.08 \com{0.39} & 0.74 \com{0.02} & 78 & 101 & 58 & 2955 & 1011 & \cellcolor{No} 2.11 \com{0.04} & \cellcolor{No} 4.02 \com{0.16} \\
LLaMA-3.1-8B & 0.89 \com{0.02} & 10.84 \com{0.18} & 12.41 \com{0.31} & 0.74 \com{0.02} & 1933 & 381 & 1017 & 545 & 326 & \cellcolor{No} 1.54 \com{0.22} & \cellcolor{No} 3.79 \com{1.02} \\
Mistral-7B & 0.83 \com{0.03} & 14.46 \com{0.28} & 1.10 \com{0.02} & 0.64 \com{0.03} & 1519 & 2684 & 0 & 0 & 0 & \cellcolor{Yes} 0.00 \com{0.00} & \cellcolor{Yes} 0.00 \com{0.00} \\
FLAN-T5-large & 1.01 \com{0.04} & 24.45 \com{0.28} & 47.91 \com{0.55} & 0.68 \com{0.04} & 0 & 0 & 0 & 0 & 4203 & \cellcolor{No} 4.90 \com{0.01} & \cellcolor{No} 11.14 \com{0.36} \\
FLAN-T5-xl & 0.85 \com{0.05} & 12.95 \com{0.15} & 9.59 \com{0.27} & 0.65 \com{0.05} & 1452 & 1053 & 1408 & 0 & 289 & \cellcolor{Yes} 0.00 \com{0.00} & \cellcolor{Yes} 0.00 \com{0.00} \\
Gemma-2-2B & 0.74 \com{0.04} & 0.04 \com{0.00} & 0.00 \com{0.00} & 0.74 \com{0.04} & 4203 & 0 & 0 & 0 & 0 & \cellcolor{Yes} 0.00 \com{0.00} & \cellcolor{Yes} 0.00 \com{0.00} \\
Gemma-2-9B & 0.74 \com{0.04} & 0.04 \com{0.00} & 0.00 \com{0.00} & 0.74 \com{0.04} & 4203 & 0 & 0 & 0 & 0 & \cellcolor{Yes} 0.00 \com{0.00} & \cellcolor{Yes} 0.00 \com{0.00} \\
GPT-3.5-turbo & 0.99 \com{0.02} & 19.96 \com{0.51} & 33.75 \com{0.66} & 0.73 \com{0.01} & 527 & 190 & 504 & 692 & 2292 & \cellcolor{No} 2.97 \com{0.15} & \cellcolor{No} 5.86 \com{0.39} \\
GPT-4o-mini & 0.74 \com{0.04} & 0.04 \com{0.00} & 0.00 \com{0.00} & 0.74 \com{0.04} & 4203 & 0 & 0 & 0 & 0 & \cellcolor{Yes} 0.00 \com{0.00} & \cellcolor{Yes} 0.00 \com{0.00} \\
GPT-4o & 0.85 \com{0.05} & 12.92 \com{0.29} & 0.98 \com{0.02} & 0.67 \com{0.05} & 1807 & 2396 & 0 & 0 & 0 & \cellcolor{No} 2.26 \com{0.01} & \cellcolor{Yes} 0.00 \com{0.00} \\
OpenAI o1-mini \textsuperscript{*} & 0.77 \com{0.03} & 8.21 \com{0.43} & 0.62 \com{0.03} & 0.66 \com{0.03} & 222 & 127 & 0 & 0 & 0 & \cellcolor{No} 1.13 \com{0.81} & \cellcolor{Yes} 0.00 \com{0.00} \\
OpenAI o1 \textsuperscript{*} & 0.76 \com{0.02} & 5.18 \com{0.16} & 0.39 \com{0.01} & 0.69 \com{0.02} & 269 & 80 & 0 & 0 & 0 & \cellcolor{No} 0.38 \com{0.54} & \cellcolor{Yes} 0.00 \com{0.00} \\
OpenAI o3-mini \textsuperscript{*} & 0.86 \com{0.04} & 16.91 \com{1.18} & 1.28 \com{0.09} & 0.63 \com{0.03} & 87 & 262 & 0 & 0 & 0 & \cellcolor{No} 2.09 \com{0.06} & \cellcolor{Yes} 0.00 \com{0.00} \\
\bottomrule

\multicolumn{11}{c}{\textbf{\centering \rule{0pt}{10pt} Comparison with SOTA Exploration-Decision Baselines}} \\ 
\toprule
AIwRG \cite{he2024large} & 1.02 \com{0.05} & 23.73 \com{0.98} & 44.01 \com{3.50} & 0.69 \com{0.06} & 113 & 244 & 0 & 0 & 3852 & \cellcolor{No} 4.68 \com{0.18} & \cellcolor{No} 9.26 \com{1.83} \\
PerLLM \cite{yang2024perllm} & 0.97 \com{0.06} & 21.12 \com{0.08} & 1.60 \com{0.01} & 0.66 \com{0.06} & 281 & 3922 & 0 & 0 & 0 & \cellcolor{Yes} 0.00 \com{0.00} & \cellcolor{Yes} 0.00 \com{0.00} \\

\Name (DQN) & 1.07 \com{0.02} & 18.42 \com{1.17} & 21.39 \com{2.40} & 0.81 \com{0.02} & 133 & 370 & 2291 & 1153 & 230 & \cellcolor{No} 4.64 \com{0.44} & \cellcolor{No} 3.61 \com{1.20} \\
\Name (RC-DQN) & 0.98 \com{0.09} & 11.04 \com{1.88} & 11.90 \com{1.78} & 0.82 \com{0.07} & 1451 & 15 & 2572 & 114 & 24 & \cellcolor{No} 0.51 \com{0.36} & \cellcolor{Yes} 0.00 \com{0.00} \\
\Name (PPO) & 1.04 \com{0.06} & 16.83 \com{1.21} & 19.18 \com{2.79} & 0.80 \com{0.04} & 247 & 119 & 2971 & 803 & 38 & \cellcolor{No} 3.36 \com{0.16} & \cellcolor{No} 0.86 \com{1.21} \\
\Name (RC-PPO) & 1.01 \com{0.05} & 13.08 \com{0.29} & 13.72 \com{0.30} & 0.81 \com{0.04} & 884 & 0 & 3293 & 0 & 0 & \cellcolor{Yes} 0.00 \com{0.00} & \cellcolor{Yes} 0.00 \com{0.00} \\
\Name (A2C) & 1.08 \com{0.06} & 16.69 \com{0.15} & 17.75 \com{0.31} & 0.84 \com{0.06} & 0 & 0 & 4083 & 93 & 0 & \cellcolor{No} 4.21 \com{0.62} & \cellcolor{Yes} 0.00 \com{0.00} \\
\rowcolor{Yes} \textbf{\Name (RC-A2C)} & \textbf{1.06 \com{0.04}} & \textbf{13.07 \com{1.81}} & \textbf{13.71 \com{1.91}} & \textbf{0.86 \com{0.02}} & \textbf{886} & \textbf{0} & \textbf{3291} & \textbf{0} & \textbf{0} & \textbf{\cellcolor{Yes} 0.00 \com{0.00}} & \textbf{\cellcolor{Yes} 0.00 \com{0.00}} \\

\bottomrule
\end{tabular}
}

\vspace{2pt}
\justifying

\noindent{\footnotesize \textsuperscript{*} Due to the considerable inference time and usage cost of OpenAI o1-mini, OpenAI o1, and OpenAI o3-mini, we sample 100 instances for evaluation.}
\vspace{-2mm}
\end{table*}

\noindent\textbf{Baselines.} We compare the RL policies with three naive baselines: Random (with random LLM and modality selection), On-device (using only the on-device LLM), and Cloud (using only the cloud LLM with random modality selection). Inspired by recent advances in LLMs for routing \cite{ding2024hybrid, fang2025collaborative}, we substitute RL with LLM-as-Router as \Name's decision engine for the selection of LLMs and modalities. Specifically, current states and possible actions are translated into natural language descriptions as contextual input for the LLM routers, which then output an action in response to this prompt, following an RL-like procedure. The LLM routers include Phi-3-mini, Phi-3.5-mini \cite{abdin2024phi}, LLaMA-3.2-3B, LLaMA-3.1-8B \cite{llama3}, Mistral-7B \cite{jiang2023mistral}, FLAN-T5-large, FLAN-T5-xl \cite{chung2024scaling}, Gemma-2-2B, Gemma-2-9B \cite{team2024gemma2}, GPT-3.5-turbo, GPT-4o-mini, GPT-4o, OpenAI o1-mini, OpenAI o1, and OpenAI o3-mini \cite{OpenAI}. We reiterate that none of the existing works can be directly applied to our setting, which features multi-modal, multi-task, and multi-turn characteristics. Therefore, we reimplement AIwRG \cite{he2024large}, an active inference method, and PerLLM \cite{yang2024perllm}, a bandit-based approach using upper confidence bound (UCB) for exploration-exploitation trade-off, integrating the settings introduced in this paper to adapt them to our research problem.

\noindent\textbf{Default Settings.} The default experimental settings are as follows unless otherwise stated; we also study the effects of various system parameters. Specifically, we use RC-A2C as the RL policy, with $\alpha=1$, $\beta_{\Lambda} = 1/3$, $\beta_\psi = 1/3$, and $\beta_\phi = 1/3$ for the reward function weights, a 30 second latency constraint, a 0.05 USD usage cost constraint, and Jetson TX2 as the device type. These settings serve as the baseline from which variations are systematically introduced to assess their effects. Additionally, we set the time span $\tau=5$, $k=5$ nearest neighbors for response score estimation, and $\kappa = 4.63 \times 10^{-8}$ (USD per Joule) based on the average electricity price in the US \footnote{\url{https://www.energybot.com/electricity-rates/}}.

\subsection{Main Experimental Results}

Table \ref{Table Main Result} presents the main results of this study, with key takeaways described from four perspectives. 

\noindent\textbf{Comparison with Naive Baselines.} Compared to the on-device baseline, the proposed methods can significantly improve the response score at the expense of increased latency and usage cost, thus enhancing the overall reward. Compared to the cloud baseline, we also boost the response score and conserve resources by selecting the appropriate modalities, which leads to improved rewards. In contrast to the Random baseline, our approach strategically explores the relationships between tasks/conversation turns and LLMs/modalities, thereby optimizing decision-making processes to better align with the specific requirements and constraints of each task.

\noindent\textbf{Comparison with SOTA LLM-as-Router Baselines.} We replace the RL policy in \Name with an LLM router for the decision-making process. Our goal is to evaluate the zero-shot LLM router's ability to consider task-modality associations, understand contexts, and handle numerically sensitive tasks (e.g., avoiding resource constraint violations). As shown in Table \ref{Table Main Result}, different LLM routers exhibit substantial variability in performance. Some LLM routers ignore all modalities (e.g., Mistral-7B, Gemma-2-2b, Gemma-2-9b, GPT-4o-mini, GPT-4o, OpenAI o1-mini, OpenAI o1, and OpenAI o3-mini), while others choose to upload all available modalities for each task (e.g., FLAN-T5-large). Although some LLM routers (e.g., Phi-3-mini, Phi-3.5-mini, and FLAN-T5-xl) appear to select different LLMs and modalities without violating any constraints, their response scores and overall rewards do not confer significant advantages. Thus, these zero-shot LLM routers do not demonstrate clear benefits. This is unsurprising given the inherent complexity of the routing problem. As discussed in Sec. \ref{Sec. Challenges}, the decision-making process involves navigating trade-offs between multi-modal redundancy, resource constraints, and cumulative costs across conversation turns -- challenges that even humans find difficult to optimize. Without exposure to the specific task dynamics, zero-shot LLMs lack the nuanced understanding required for effective routing. Consequently, some routers violate constraints, whereas others adopt overly conservative strategies.

\newcommand{\cmark}{\ding{51}}
\newcommand{\xmark}{\ding{55}}

\begin{table*}[t]
\caption{Ablation Study. Using RC-A2C as backbone.}
\vspace{-2mm}
\label{Table Ablation}
\centering
\begin{tabular}{
>{\centering\arraybackslash}p{14mm}
>{\centering\arraybackslash}p{14mm}
>{\centering\arraybackslash}p{14mm}
>{\centering\arraybackslash}p{14mm}
|ccc|a|cc
}
\toprule
LLM & Modality & Resource & Resp. Score & \textbf{Response} & \textbf{Latency} & \textbf{Usage Cost} & \textbf{Overall} & \multicolumn{2}{c}{\textbf{Constraint Violation} ($\downarrow$)} \\
Selection & Selection & Awareness & Estimation & \textbf{Score} ($\uparrow$) & (s) ($\downarrow$) & (1e-3 USD) ($\downarrow$) & \textbf{Reward} ($\uparrow$) & Latency & Usage Cost \\
\midrule
\xmark &  &  &  & 0.86 \com{0.06} & 7.55 \com{0.25} & 7.90 \com{0.26} & 0.76 \com{0.06} & \cellcolor{Yes} 0.00 \com{0.00} & \cellcolor{Yes} 0.00 \com{0.00} \\
& \xmark &  &  & 0.99 \com{0.01} & 15.96 \com{0.13} & 19.76 \com{0.11} & 0.75 \com{0.01} & \cellcolor{No} 0.93 \com{0.42} & \cellcolor{No} 1.83 \com{2.60} \\
&  & \xmark &  & 0.90 \com{0.04} & 9.48 \com{2.93} & 9.93 \com{3.08} & 0.76 \com{0.02} & \cellcolor{Yes} 0.00 \com{0.00} & \cellcolor{Yes} 0.00 \com{0.00} \\
&  &  & \xmark & 1.00 \com{0.02} & 14.19 \com{0.89} & 14.90 \com{0.93} & 0.78 \com{0.03} & \cellcolor{Yes} 0.00 \com{0.00} & \cellcolor{Yes} 0.00 \com{0.00} \\
\rowcolor{Yes} \cmark & \cmark & \cmark & \cmark & \textbf{1.06 \com{0.04}} & \textbf{13.07 \com{1.81}} & \textbf{13.71 \com{1.91}} & \textbf{0.86 \com{0.02}} & \textbf{\cellcolor{Yes} 0.00 \com{0.00}} & \textbf{\cellcolor{Yes} 0.00 \com{0.00}} \\

\bottomrule
\end{tabular}
\vspace{-4mm}
\end{table*}

\noindent\textbf{Comparison with SOTA Exploration-Decision Baselines.} Existing works do not consider multi-task and multi-turn settings in multi-modal scenarios, and cannot be trivially deployed to our setting. For example, AIwRG~\cite{he2024large} utilizes a benchmark with a fixed evaluation metric Pass@k that measures success rates. In contrast, our method addresses multi-modal, multi-task, and multi-turn conversation settings, which introduce significant challenges related to response score evaluation and uncertainty. From the results, our \Name system significantly outperforms these SOTA baselines. \mbox{AIwRG}~\cite{he2024large} fails to efficiently select LLMs and modalities, frequently opting to upload all three types of modality data sources for most actions. This results in substantial resource waste and redundancy in uploading unnecessary information to the cloud. Conversely, PerLLM~\cite{yang2024perllm} rarely considers any modality data, relying solely on the on-device LLM and text-only cloud LLM. Although this approach avoids violating any resource constraints, it leads to lower response scores, as the cloud LLM lacks access to information derived from modality data.

\noindent\textbf{Analysis of Our \Name System.} Most RL methods prefer utilizing the cloud LLM over the on-device LLM, as the gain in response score from the cloud LLM substantially outweighs the resource savings achieved by the on-device LLM. Regarding modality selection, few RL methods choose text-only cloud LLM queries, as they often result in interaction times comparable to or even longer than those with multi-modal inputs. This phenomenon depends on the inference speed of the service provider rather than factors such as network bandwidth. A potential explanation is that text-only and multi-modal cloud LLMs may operate on different workflows or model versions. However, for cases involving multi-modal inputs, we observe that single-modality selection is often preferred by RL methods, as it provides valuable information with lower latency and reduced usage costs compared to selecting two or three modalities. RC-A2C demonstrates the best response scores and overall rewards while maintaining compliance with latency and usage cost constraints.

\newcolumntype{C}[1]{>{\centering\arraybackslash}p{#1}}

\begin{table}[t]
\caption{Trade-off between Metrics in Reward Function - Varying weights illustrate preferences for different metrics.}
\label{Table Trade-off}
\vspace{-2mm}
\centering 
\resizebox{\linewidth}{!}{
\begin{tabular}{c|l|cccc|a}
\toprule
\multirow{2}{*}{$(\beta_{\Lambda},\,\beta_\psi,\,\beta_\phi)$} & \multirow{2}{*}{\textbf{Method}} & \textbf{Response} & \textbf{Association} & \textbf{Latency} & \textbf{Usage Cost} & \textbf{Overall} \\
 & & \textbf{Score} ($\uparrow$) & ($\uparrow$) & (s) ($\downarrow$) & (1e-3 USD) ($\downarrow$) & \textbf{Reward} ($\uparrow$) \\
\midrule
\multirow{3}{*}{(0,\,0.5,\,0.5)} & AIwRG & 0.99 \com{0.03} & 0.00 \com{0.00} & 22.56 \com{0.21} & 1.71 \com{0.02} & 0.51 \com{0.03} \\
 & PerLLM & 0.97 \com{0.04} & 0.18 \com{0.00} & 13.13 \com{0.04} & 13.78 \com{0.04} & 0.53 \com{0.04} \\
 & \cellcolor{Yes} \textbf{Ours} & \cellcolor{Yes} \textbf{0.71 \com{0.04}} & \cellcolor{Yes} \textbf{0.00 \com{0.00}} & \cellcolor{Yes} \textbf{0.04 \com{0.00}} & \cellcolor{Yes} \textbf{0.00 \com{0.00}} & \cellcolor{Yes} \textbf{0.71 \com{0.04}} \\
\midrule
\multirow{3}{*}{(0.5,\,0,\,0.5)} & AIwRG & 0.99 \com{0.03} & 0.00 \com{0.00} & 22.56 \com{0.21} & 1.71 \com{0.02} & 0.97 \com{0.03} \\
 & PerLLM & 0.95 \com{0.07} & 0.06 \com{0.08} & 13.01 \com{0.10} & 5.25 \com{6.05} & 0.94 \com{0.07} \\
 & \cellcolor{Yes} \textbf{Ours} & \cellcolor{Yes} \textbf{1.05 \com{0.06}} & \cellcolor{Yes} \textbf{0.18 \com{0.01}} & \cellcolor{Yes} \textbf{14.09 \com{0.91}} & \cellcolor{Yes} \textbf{14.79 \com{0.96}} & \cellcolor{Yes} \textbf{1.04 \com{0.06}} \\
\midrule
\multirow{3}{*}{(0.5,\,0.5,\,0)} & AIwRG & 1.04 \com{0.04} & 0.21 \com{0.00} & 16.63 \com{0.15} & 17.46 \com{0.16} & 0.86 \com{0.04} \\
 & \textbf{PerLLM} & \textbf{1.00 \com{0.05}} & \textbf{0.21 \com{0.04}} & \textbf{12.91 \com{0.29}} & \textbf{15.47 \com{2.42}} & \textbf{0.88 \com{0.01}} \\
 & \cellcolor{Yes} Ours & \cellcolor{Yes} 1.02 \com{0.05} & \cellcolor{Yes} 0.18 \com{0.01} & \cellcolor{Yes} 13.82 \com{0.94} & \cellcolor{Yes} 14.51 \com{0.99} & \cellcolor{Yes} 0.87 \com{0.06} \\
\midrule
\multirow{3}{*}{(1/3,\,1/3,\,1/3)} & AIwRG & 0.99 \com{0.03} & 0.00 \com{0.00} & 22.56 \com{0.21} & 1.71 \com{0.02} & 0.67 \com{0.03} \\
 & PerLLM & 1.01 \com{0.10} & 0.12 \com{0.08} & 13.08 \com{0.10} & 9.53 \com{6.05} & 0.81 \com{0.10} \\
 & \cellcolor{Yes} \textbf{Ours} & \cellcolor{Yes} \textbf{1.06 \com{0.04}} & \cellcolor{Yes} \textbf{0.17 \com{0.02}} & \cellcolor{Yes} \textbf{13.07 \com{1.81}} & \cellcolor{Yes} \textbf{13.71 \com{1.91}} & \cellcolor{Yes} \textbf{0.86 \com{0.02}} \\
\bottomrule
\end{tabular}
}
\vspace{-2mm}
\end{table}

\subsection{Ablation Study}

To validate our approach, we conduct ablation studies in Table \ref{Table Ablation} by systematically removing key components to address the following research questions: \textbf{RQ1}: What is the impact of LLM selection? \textbf{RQ2}: How does modality selection affect performance? \textbf{RQ3}: Is resource awareness necessary? \textbf{RQ4}: Does uncertain response score estimation improve decision-making? Specifically, we evaluate variants where: (i) LLM selection is replaced with random selection, (ii) modality selection is replaced with random selection, (iii) resource awareness is removed by employing the original state space $\mathcal{S}$ instead of the resource-aware $\mathcal{S}^\text{RA}$, and (iv) uncertain response score estimation is replaced with average action scores. Our findings reveal several critical insights. Random LLM selection significantly degrades response scores, as on-device LLMs frequently lack the requisite capabilities for certain tasks. Similarly, random modality selection fails to optimize modality-task alignment, resulting in suboptimal performance. The absence of resource awareness leads to overly conservative policies, as the RL policy cannot effectively reason about available resources, and consequently underutilizes resource budgets. Lastly, using average response scores fails to account for contextual information, particularly historical modality uploads, leading to biased estimations and degraded performance. These results collectively validate the necessity of each component in achieving optimal performance across all evaluation metrics.

\subsection{Trade-off Between Metrics}

Table~\ref{Table Trade-off} compares \Name against AIwRG and PerLLM across four reward weight configurations, obtained by sweeping $\beta_{\Lambda}$, $\beta_\psi$, and $\beta_\phi$ for association, latency, and usage cost in Eq.~\eqref{eq:reward} with $\alpha=1$ fixed. The baselines exhibit clear failure modes across weight configurations: AIwRG collapses to an on-device-only policy in three of the four settings, producing nearly identical metrics regardless of $(\beta_\psi,\beta_\phi)$ and failing to adapt to the specified preferences, while PerLLM does not consistently track the reward weights either, for instance, under $(0,\,0.5,\,0.5)$ it still incurs a substantial latency and usage cost despite the weighting explicitly penalizing it. In contrast, \Name attains the highest overall reward in three of four configurations and is statistically tied in the fourth, confirming that its advantage holds across the diverse application scenarios induced by different weightings (e.g.,\ cost-sensitive vs.\ latency-sensitive deployments) rather than being an artifact of any single weighting.

\begin{figure}[t]
    \centering
    \subfloat[Latency Constraint]{\includegraphics[width=0.5\linewidth]{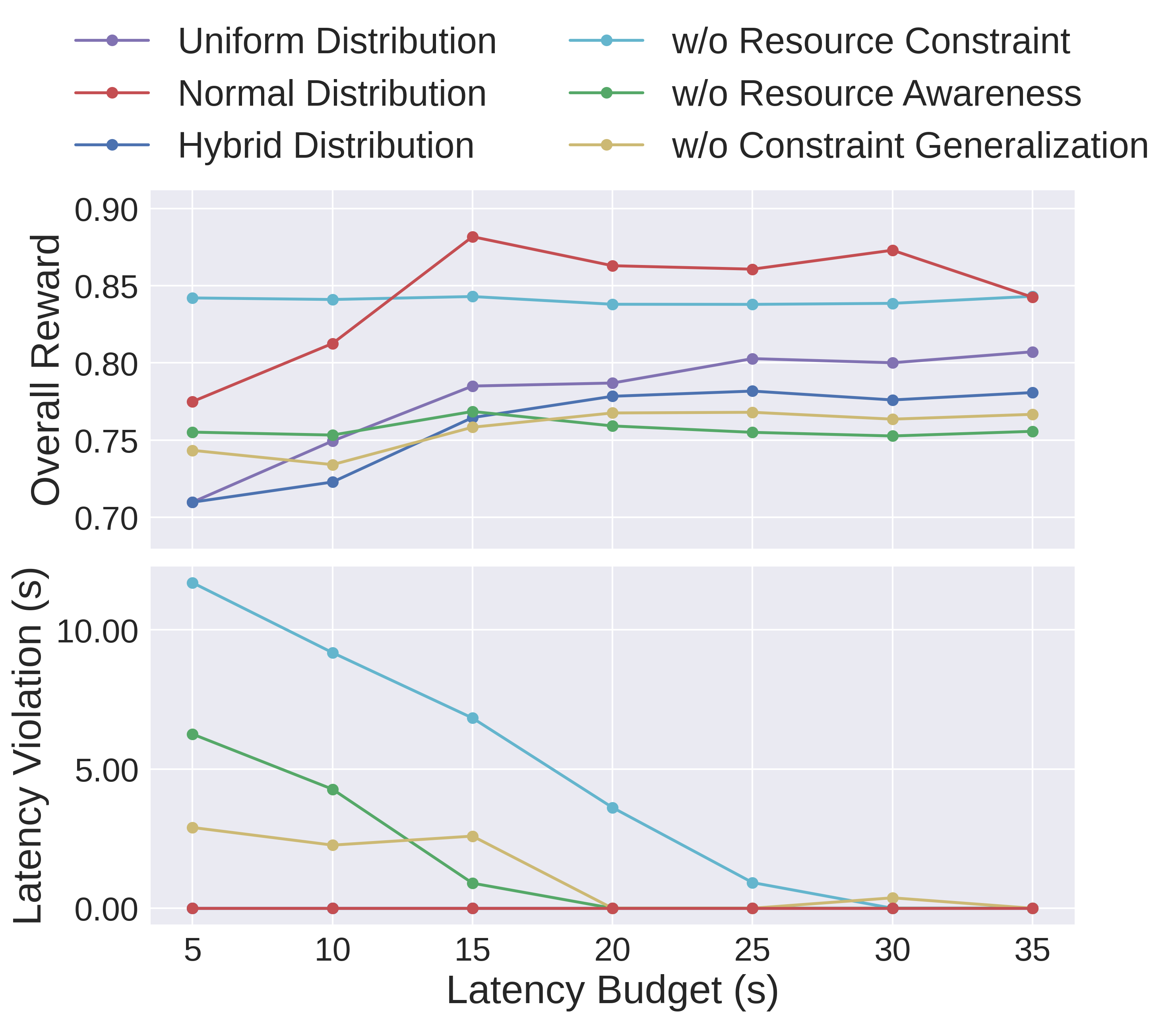}}
    \hfill
    \subfloat[Usage Cost Constraint]{\includegraphics[width=0.5\linewidth]{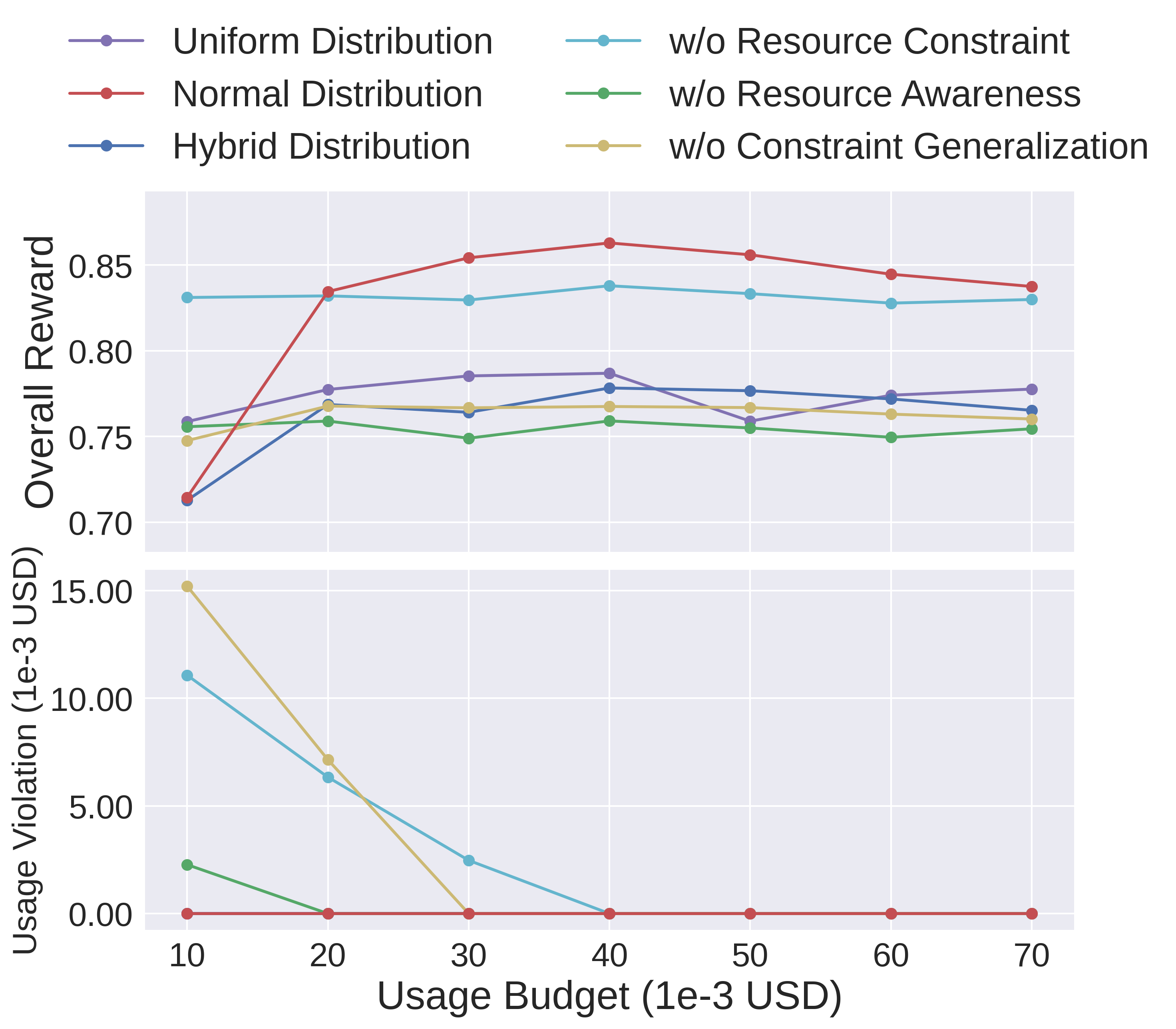}}
    \caption{Effect of resource constraints. \Name maintains compliance with both latency and usage cost constraints under different post-training budget configurations, outperforming baseline methods and demonstrating robust generalization to diverse user-specified resource preferences.}
    \label{Fig. Effect Constraints}
\vspace{-4mm}
\end{figure}

\begin{table}[t]
\caption{On-Device Hardware Specifications.}
\vspace{-2mm}
\label{Table Device}
\centering
\resizebox{\linewidth}{!}{
\begin{tabular}{lcccc}
\toprule
\multirow{2}{*}{\textbf{Hardware Platform}} & \textbf{Performance} & \textbf{Power} & \textbf{Latency} & \textbf{Usage Cost} \\ 
& (TFLOPS) & (Watts) & (s) & (1e-3 USD) \\ 
\midrule
Raspberry Pi-4B & 0.0135 & 8 & 1.12593 & 4.17e-4 \\
Raspberry Pi-5 & 0.0314 & 12 & 0.48408 & 2.69e-4 \\
Jetson Nano & 0.472 & 10 & 0.03220 & 1.49e-5 \\
Jetson TX2 & 1.33 & 15 & 0.01143 & 7.94e-6 \\
Jetson Xavier NX & 21 & 20 & 0.00072 & 6.71e-7 \\
Jetson Orin NX & 100 & 25 & 0.00015 & 1.76e-7 \\
iPhone 15 (A16) & 15.8 & 15 & 0.00096 & 6.69e-7 \\
iPhone 15 Pro (A17 Pro) & 35 & 15 & 0.00043 & 3.02e-7 \\
\bottomrule
\end{tabular}
}
\vspace{-4mm}
\end{table}

\begin{figure}[t]
    \centering
    \subfloat[Response Score]{\includegraphics[width=0.5\linewidth]{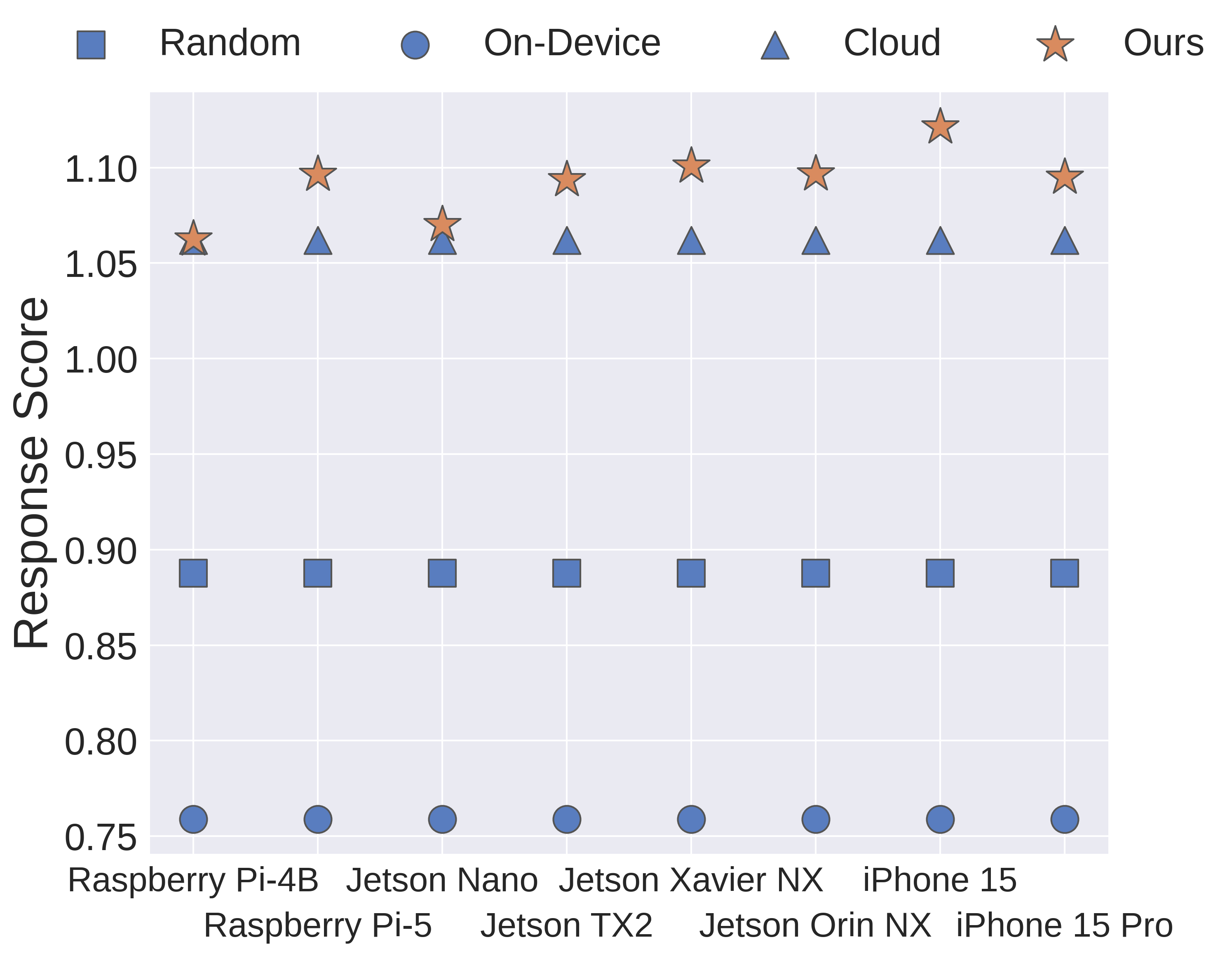}}
    \hfill
    \subfloat[Overall Reward]{\includegraphics[width=0.5\linewidth]{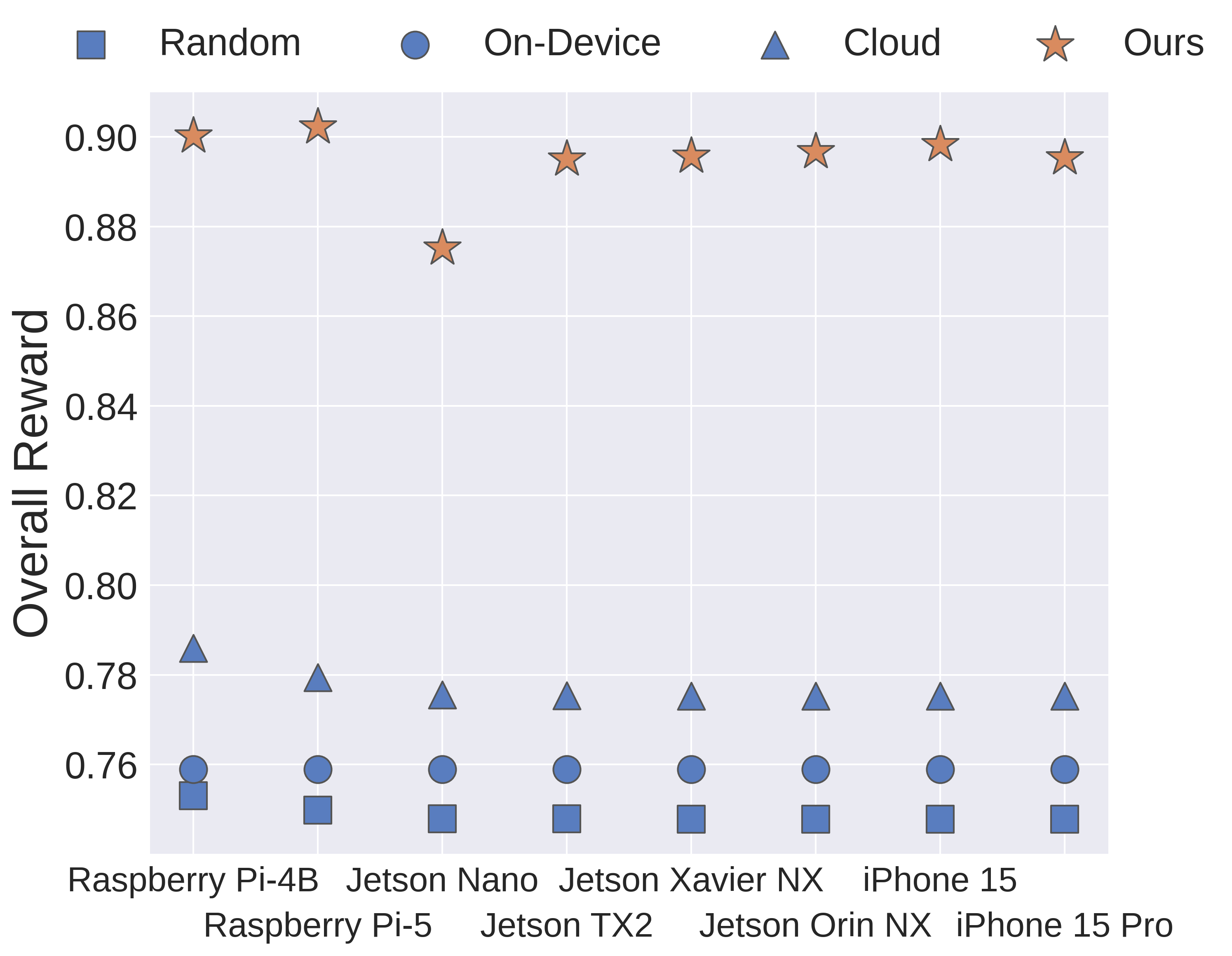}}
    \caption{Effect of user device performance. \Name achieves superior performance across all device platforms, demonstrating insensitivity to hardware variations.}
    \label{Fig. Effect Device}
\vspace{-4mm}
\end{figure}

\begin{figure}[t]
    \centering
    \subfloat[Response Score]{\includegraphics[width=0.5\linewidth]{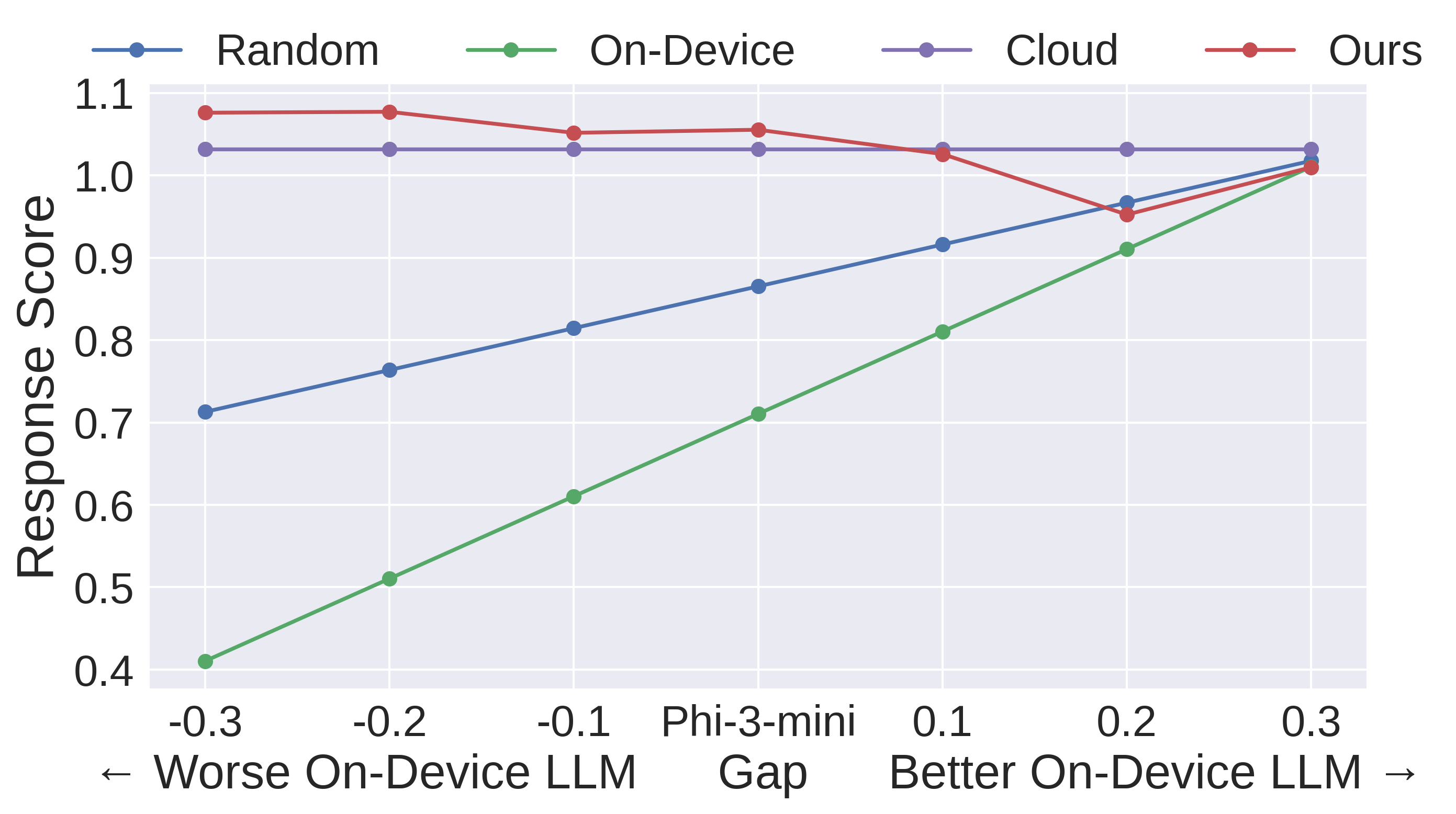}}
    \hfill
    \subfloat[Overall Reward]{\includegraphics[width=0.5\linewidth]{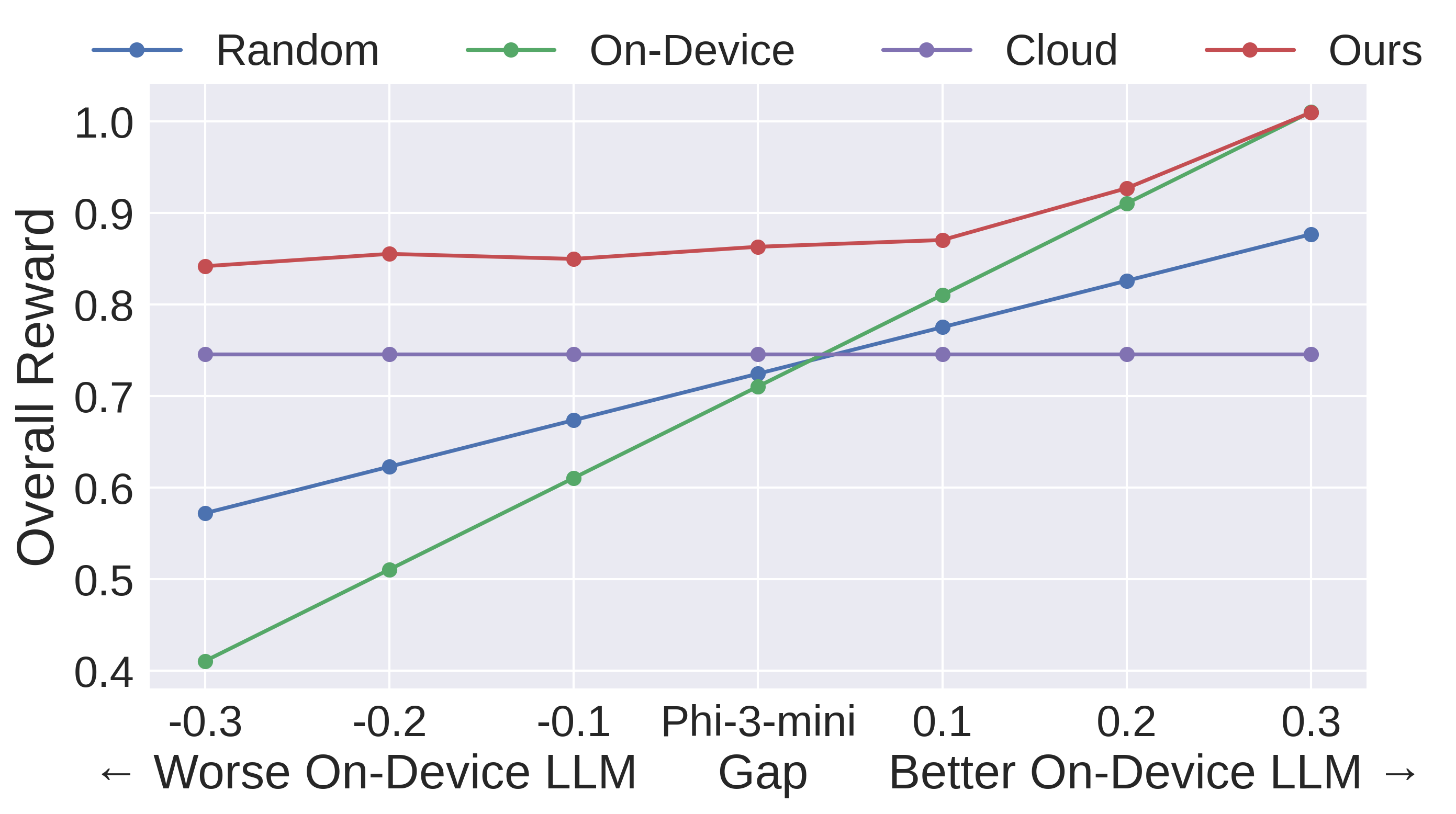}}
    \caption{Effect of response quality of on-device LLM. \Name outperforms baselines across various deployments of on-device and cloud LLMs.}
    \label{Fig. Effect LLM Gap}
\vspace{-4mm}
\end{figure}

\subsection{Effect of Resource Constraints}

We introduce the key idea of resource constraints in Sec.~\ref{Sec. Resource Awareness}, which aims to enhance the RCRL's awareness of resource constraints and its generalization capability by randomizing resource budgets during training. Specifically, the initial resource budget for each episode is sampled from a distribution $\mathcal{B}$. 
Formally, for each resource type $j \in \mathcal{J}$, we define the following budget distributions: (i) \textit{Uniform distribution}: $\xi_j \sim \mathcal{U}_j$ provides unbiased coverage across the budget range; (ii) \textit{Normal distribution}: $\xi_j \sim \mathcal{N}_j$ concentrates sampling around typical operating budgets; and (iii) \textit{Hybrid distribution}: we construct a hybrid distribution $\mathcal{H}_j$ as $\xi_j \sim \mathcal{H}_j = 0.5 \cdot \mathcal{N}_j + 0.3 \cdot \mathcal{U}_j + 0.2 \cdot \mathcal{E}_j$, where $\mathcal{E}_j$ represents a distribution concentrated on extreme budget edge cases (e.g., very low or very high budgets). To validate the effectiveness of our resource-aware design, we compare against three ablation baselines: (i) \textit{w/o Resource Constraint}: the policy is trained without any resource constraints; (ii) \textit{w/o Resource Awareness}: the policy uses the original state space $\mathcal{S}$ instead of the resource-aware state space $\mathcal{S}^\text{RA}$; and (iii) \textit{w/o Constraint Generalization}: the policy is trained with fixed resource budgets rather than randomized distributions. During evaluation, all methods are tested under identical post-training budget configurations to assess their generalization capabilities.

We evaluate the impact of these distribution strategies and baseline comparisons on RCRL training performance in Fig. \ref{Fig. Effect Constraints}. First, among the distribution strategies, the normal distribution consistently outperforms both uniform and hybrid distributions across all evaluation metrics. This superiority can be attributed to the fact that uniform and hybrid distributions expose the RL policy to an excessive proportion of edge cases during training, leading to overly conservative policies. Since the reward is set to zero whenever resource constraints are violated, the RL policy receives limited informative feedback in these extreme scenarios, hindering its ability to learn meaningful trade-offs between response quality and resource consumption. In contrast, the normal distribution provides moderately constrained budgets that necessitate careful resource management while still allowing sufficient flexibility for diverse actions. Second, comparing with baseline methods, we observe that \textit{w/o Resource Constraint} suffers from severe constraint violations under tight budgets. The \textit{w/o Resource Awareness} baseline, lacking explicit budget information in the state space, leads the policy to adopt overly conservative actions, resulting in lower overall reward. Similarly, \textit{w/o Constraint Generalization} exhibits poor performance under lower budgets, indicating insufficient robustness to diverse budget configurations. In contrast, our method with normal distribution achieves near-zero violations across all budget settings while maintaining competitive reward performance, validating the effectiveness of combining resource-aware state representation with constraint generalization.

\subsection{Effect of User Device}
To further validate the robustness of our approach, we investigate the effects of deploying \Name on different types of user devices in Fig. \ref{Fig. Effect Device}, with detailed on-device hardware specifications provided in Table \ref{Table Device}. Our results demonstrate that the performance variations across different devices do not significantly impact the operation of the \Name system, as evidenced by the absence of statistically significant differences in response scores and overall rewards. The underlying rationale is that for any device, latency and usage cost are normalized through min-max scaling and incorporated into the overall reward formulation, ensuring consistent evaluation across heterogeneous hardware configurations. Through strategic LLM and modality selection, our approach achieves response quality comparable to or marginally surpassing that of the cloud LLM baseline, while simultaneously incurring substantially lower latency and usage cost, thereby yielding significantly superior overall rewards. Consequently, given that the latency and usage cost of on-device LLMs are inherently lower than those of cloud LLMs, a key takeaway is that the \Name system can operate effectively across diverse device platforms without significant performance degradation.

\begin{table*}[t]
\centering
\caption{Cloud Latency Sensitivity - Without retraining, \Name shifts traffic from cloud to device as latency degrades, with zero usage cost violations across all scaling factors and latency violations only when a single cloud call alone exceeds the budget.}
\vspace{-2mm}
\label{Table Cloud Latency Sensitivity}
\resizebox{\linewidth}{!}{
\begin{tabular}{c|ccc|a|ccccc|cc}
\toprule
\textbf{Cloud Latency} & \textbf{Response} & \textbf{Latency} & \textbf{Usage Cost} & \textbf{Overall} & \multirow{2}{*}{\textbf{Device}} & \multicolumn{4}{c|}{\textbf{Cloud - Num. Selected Modalities}} & \multicolumn{2}{c}{\textbf{Constraint Violation} ($\downarrow$)} \\
\textbf{Multipliers} & \textbf{Score} ($\uparrow$) & (s) ($\downarrow$) & (1e-3 USD) ($\downarrow$) & \textbf{Reward} ($\uparrow$) & & 0 (text-only) & 1 & 2 & 3 & Latency & Usage Cost \\
\midrule
1$\times$ & 1.06 \com{0.04} & 13.07 \com{1.81} & 13.71 \com{1.91} & 0.86 \com{0.02} & 886 & 0 & 3291 & 0 & 0 & \cellcolor{Yes} 0.00 \com{0.00} & \cellcolor{Yes} 0.00 \com{0.00} \\
2$\times$ & 1.01 \com{0.05} & 26.12 \com{0.12} & 13.71 \com{0.07} & 0.82 \com{0.05} & 900 & 0 & 3291 & 0 & 0 & \cellcolor{Yes} 0.00 \com{0.00} & \cellcolor{Yes} 0.00 \com{0.00} \\
3$\times$ & 0.94 \com{0.05} & 28.59 \com{0.00} & 10.00 \com{0.00} & 0.79 \com{0.05} & 1791 & 0 & 2400 & 0 & 0 & \cellcolor{Yes} 0.00 \com{0.00} & \cellcolor{Yes} 0.00 \com{0.00} \\
5$\times$ & 0.83 \com{0.05} & 31.77 \com{11.22} & 6.67 \com{2.36} & 0.73 \com{0.02} & 2591 & 0 & 1600 & 0 & 0 & \cellcolor{No} 5.88 \com{8.31} & \cellcolor{Yes} 0.00 \com{0.00} \\
10$\times$ & 0.79 \com{0.00} & 47.64 \com{0.00} & 5.00 \com{0.00} & 0.71 \com{0.00} & 2991 & 0 & 1200 & 0 & 0 & \cellcolor{No} 17.64 \com{0.00} & \cellcolor{Yes} 0.00 \com{0.00} \\
\bottomrule
\end{tabular}
}
\vspace{-4mm}
\end{table*}

\begin{table}[t]
    \centering
    \caption{Effect of task distribution shift. \Name maintains its overall reward advantage across all task-concentration levels $\alpha$.}
    \vspace{-2mm}
    \label{Table Task Distribution Shift}
    \resizebox{\linewidth}{!}{
    \begin{tabular}{c|l|ccc|a}
    \toprule
    \multirow{2}{*}{$\alpha$} & \multirow{2}{*}{\textbf{Method}} & \textbf{Response} & \textbf{Latency} & \textbf{Usage Cost} & \textbf{Overall} \\
     & & \textbf{Score} ($\uparrow$) & (s) ($\downarrow$) & (1e-3 USD) ($\downarrow$) & \textbf{Reward} ($\uparrow$) \\
    \midrule
    \multirow{3}{*}{0.1} & AIwRG & 0.98 \com{0.08} & 22.22 \com{1.32} & 38.51 \com{9.56} & 0.66 \com{0.06} \\
      & PerLLM & 0.99 \com{0.05} & 21.58 \com{0.38} & 40.43 \com{0.55} & 0.68 \com{0.05} \\
      & \cellcolor{Yes} \textbf{Ours} & \cellcolor{Yes} \textbf{1.05 \com{0.04}} & \cellcolor{Yes} \textbf{13.88 \com{0.75}} & \cellcolor{Yes} \textbf{14.57 \com{0.79}} & \cellcolor{Yes} \textbf{0.84 \com{0.03}} \\
    \midrule
    \multirow{3}{*}{0.5} & AIwRG & 0.99 \com{0.06} & 25.42 \com{2.21} & 43.80 \com{12.68} & 0.68 \com{0.04} \\
      & PerLLM & 0.99 \com{0.04} & 23.55 \com{0.15} & 43.34 \com{0.25} & 0.69 \com{0.04} \\
      & \cellcolor{Yes} \textbf{Ours} & \cellcolor{Yes} \textbf{1.04 \com{0.04}} & \cellcolor{Yes} \textbf{14.97 \com{0.70}} & \cellcolor{Yes} \textbf{15.71 \com{0.73}} & \cellcolor{Yes} \textbf{0.85 \com{0.04}} \\
    \midrule
    \multirow{3}{*}{1.0} & AIwRG & 0.98 \com{0.04} & 25.58 \com{2.35} & 44.15 \com{12.90} & 0.66 \com{0.02} \\
      & PerLLM & 0.97 \com{0.02} & 23.41 \com{0.05} & 42.87 \com{0.07} & 0.67 \com{0.02} \\
      & \cellcolor{Yes} \textbf{Ours} & \cellcolor{Yes} \textbf{1.03 \com{0.06}} & \cellcolor{Yes} \textbf{14.90 \com{0.54}} & \cellcolor{Yes} \textbf{15.64 \com{0.56}} & \cellcolor{Yes} \textbf{0.84 \com{0.05}} \\
    \midrule
    \multirow{3}{*}{5.0} & AIwRG & 0.99 \com{0.04} & 21.34 \com{1.54} & 37.18 \com{9.47} & 0.67 \com{0.03} \\
      & PerLLM & 0.98 \com{0.04} & 19.95 \com{0.11} & 37.51 \com{0.20} & 0.67 \com{0.04} \\
      & \cellcolor{Yes} \textbf{Ours} & \cellcolor{Yes} \textbf{1.09 \com{0.04}} & \cellcolor{Yes} \textbf{13.06 \com{0.42}} & \cellcolor{Yes} \textbf{13.71 \com{0.45}} & \cellcolor{Yes} \textbf{0.88 \com{0.03}} \\
    \bottomrule
    \end{tabular}
    }
\vspace{-4mm}
\end{table}

\subsection{Effect of On-Device LLM Response Quality}

We evaluate the \Name system's performance by simulating varying on-device LLM response qualities. Fig. \ref{Fig. Effect LLM Gap} presents these results, where we modify the response scores in the \DatasetName dataset by applying specific adjustments (represented as the ``Gap'' in the figure) to simulate different on-device LLM capabilities. The \Name system consistently outperforms all three baselines regardless of the quality gap between on-device and cloud LLMs, demonstrating its effectiveness across arbitrary combinations of on-device and cloud LLM capabilities. When the on-device LLM response quality is substantially inferior to that of the cloud LLM, \Name consistently selects the cloud LLM and proceeds with strategic modality selection. As the quality of the on-device LLM response approaches that of the cloud LLM, \Name adaptively balances between on-device and cloud options to optimize the overall reward. A representative example is observed at Gap = 0.2, where the on-device LLM's response score closely approximates that of the cloud LLM, while its latency and usage cost remain substantially lower. Consequently, \Name exhibits a strong preference for the on-device LLM, resulting in slightly lower response scores but still higher overall rewards due to the favorable resource trade-off. Furthermore, even in the extreme scenario where identical LLMs are deployed both on-device and in the cloud, \Name still achieves superior response quality. This holds under the assumption that the on-device LLM does not support multi-modal inputs (or, from a broader perspective, the cloud LLM cannot access user multi-modal data due to privacy concerns), which reflects the distinctive challenge described in RQ1. These results validate our methodology's applicability across diverse on-device LLM capabilities and demonstrate its robustness to varying quality gaps between on-device and cloud LLMs.

\subsection{Effect of Cloud Latency Variations}

To assess robustness to runtime overheads, network drift, and provider-side changes, we scale the measured cloud latency by $\{1,2,3,5,10\}\times$ at evaluation, with RC-A2C trained once under a $30$~s latency budget. As shown in Table~\ref{Table Cloud Latency Sensitivity}, from $1\times$ to $3\times$ the policy reads the remaining budget in $s^{\text{RA}}_t$ and reroutes traffic from cloud ($3291\to2400$ single modality calls) to device ($886\to1791$ calls), maintaining zero violations despite a tripled per call latency, which confirms that $s^{\text{RA}}_t$ encodes resource availability rather than a memorized action distribution. Beyond $3\times$, latency violations are arithmetically unavoidable since a single $10\times$ cloud call ($47.64$~s) already exceeds the budget, and the observed $17.64$~s violation exactly matches the per call excess $47.64-30$, with the saturated cloud calls under $5\times$ and $10\times$ paired with rising device usage corroborating that the policy issues at most one cloud call per episode and falls back to device inference thereafter.

\begin{figure}[t]
    \centering
    \subfloat[Latency vs. Modality Count]{\includegraphics[width=0.5\linewidth]{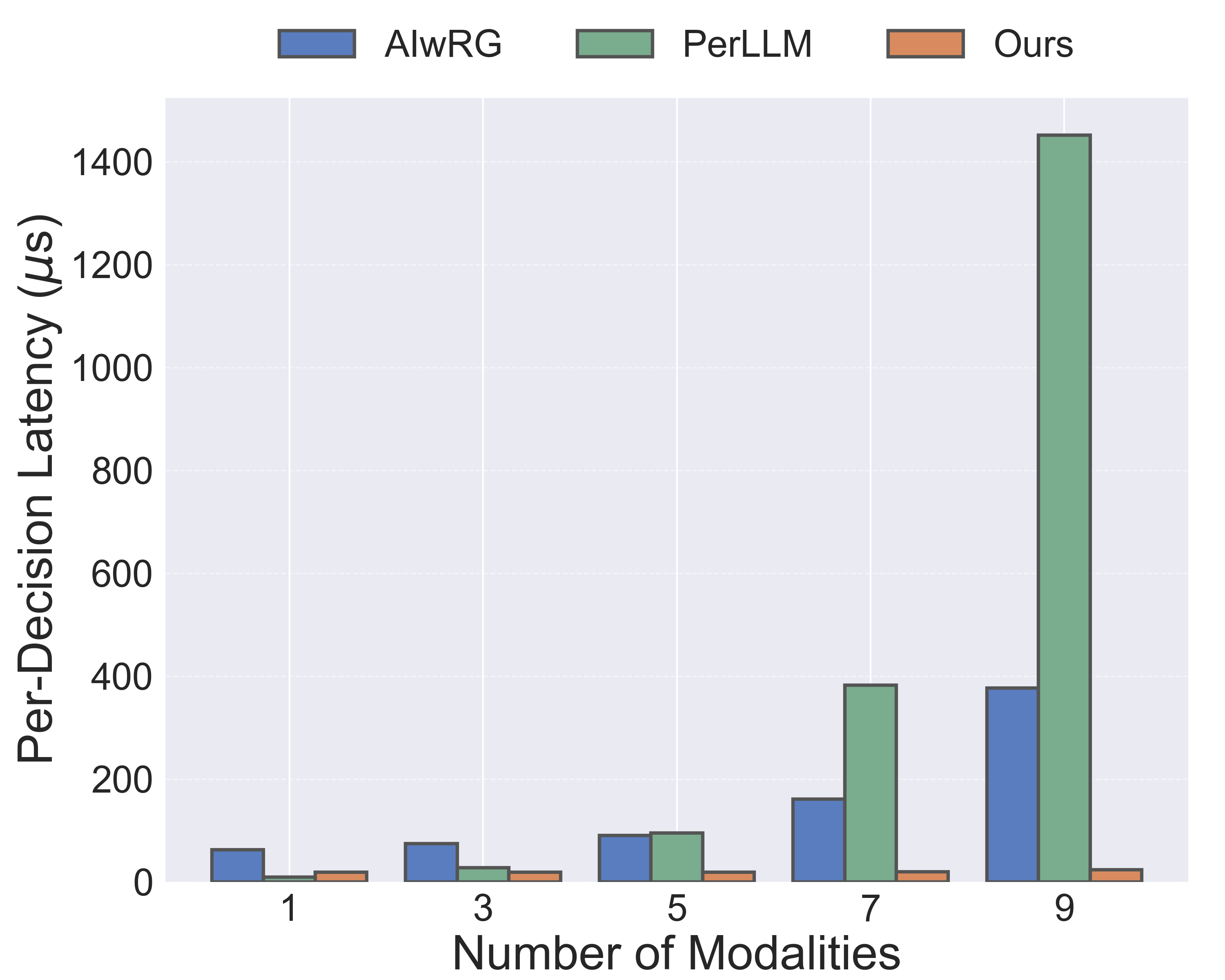}}
    \hfill
    \subfloat[Latency vs. Time Span]{\includegraphics[width=0.5\linewidth]{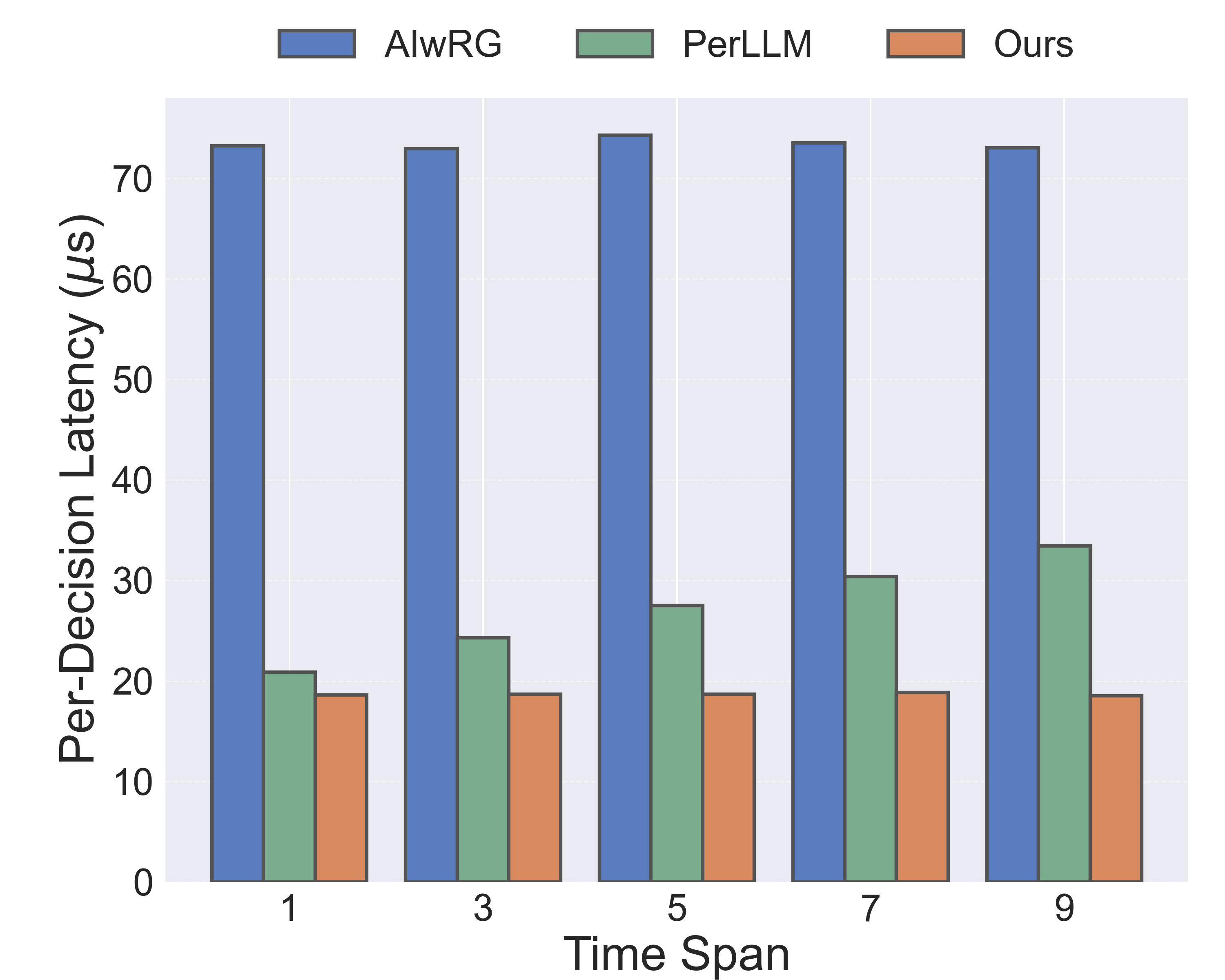}}
    \caption{Per-decision latency of the decision engine. PerLLM scales with both modality count and time span, while AIwRG scales only with modality count since time span does not enter its rollout. \Name performs a single forward pass per decision and remains essentially flat.}
    \label{Fig. Runtime}
\vspace{-4mm}
\end{figure}

\subsection{Effect of Task Distribution Shift}

Table~\ref{Table Task Distribution Shift} evaluates robustness to online task-distribution shift: each method is trained on the training set and tested on episodes whose task mixtures are drawn from $\mathrm{Dir}(\alpha \mathbf{1}_4)$, with small $\alpha$ concentrating episodes on fewer of the four categories and large $\alpha$ approaching a uniform mixture. Since \DatasetName samples categories independently per turn, episode-level concentration is absent at training time, whereas at $\alpha=0.1$ roughly $24\%$ of test episodes are dominated by a single category and over $90\%$ contain at most two, while at $\alpha=5$ over $60\%$ span three or four. For intuition, a draw at $\alpha=0.1$ gives mixture $(0.03,0.06,0.91,0.00)$, yielding a 5-turn episode (Assistant, Query, Query, Query, Query), whereas a draw at $\alpha=5$ gives mixture $(0.28,0.25,0.26,0.22)$, yielding (Edit, Assistant, Recommendation, Recommendation, Assistant). Across all $\alpha$, \Name retains its reward advantage over AIwRG and PerLLM, with all three methods varying only modestly. The limited sensitivity to $\alpha$ is consistent with the per-turn structure of the routing decision, since episode-level concentration alters the empirical mixture of turn types within an episode without changing the optimal action on any individual turn.

\subsection{Runtime Scaling Analysis}

We profile per-decision latency on an Apple M1 Pro CPU, averaged over 5{,}000 trials after 200 warmup iterations, varying the modality count $|\mathcal{M}|$ and time span $\tau$ with action-space size $|\mathcal{A}|=2^{|\mathcal{M}|}+1$, with results shown in Fig.~\ref{Fig. Runtime}. PerLLM scores every candidate placement under a UCB rule with per-arm feasibility checks, sweeping the full action set; its $\tau$-dependence arises from the sliding-window aggregation of latency/usage costs and the associated constraint-slack term, both recomputed each turn. AIwRG evaluates the expected free energy of every action in $\mathcal{A}$ via a batched forward through its belief model and takes the argmin. Both baselines therefore redo work exponential in $|\mathcal{M}|$ at every turn. \Name instead avoids enumeration: inference is a single forward pass through a fixed-architecture network of hidden width $H$ with a categorical head over $\mathcal{A}$, so $|\mathcal{M}|$ and $\tau$ affect only the input and output layer sizes. Across the tested range this growth is dominated by the constant hidden-layer cost, leaving per-decision latency essentially flat; the combinatorial cost is amortized into offline training.

\section{Conclusion}
\label{Sec. Conclusion}
 
In this paper, we developed \Name, a novel device-cloud LLM inference offloading system designed specifically for multi-modal, multi-task, and multi-turn conversation scenarios. We formulated the joint LLM and modality selection process as a resource-constrained reinforcement learning (RCRL) problem that maximizes long-term cumulative reward (response quality, latency, and usage cost) under practical user-specified resource budgets. Additionally, we curated a novel dataset, \DatasetName, enabling evaluation of \Name system performance across various configurations of modality availability, tasks, conversation turns, resource constraints, and LLM response qualities. We demonstrated improvements in the overall reward obtained by \Name over several baselines. 
Future work will extend the policy state with dynamic device- and cloud-side runtime conditions, including on-device compute, memory, and battery availability, network round-trip times and bandwidth, and provider-side load and throttling, to enable proactive rather than reactive adaptation.

\bibliographystyle{IEEEtran}
\small\bibliography{reference}

\vspace{-10mm}

\begin{IEEEbiography}
[{\includegraphics[width=1in,height=1.25in,clip,keepaspectratio]{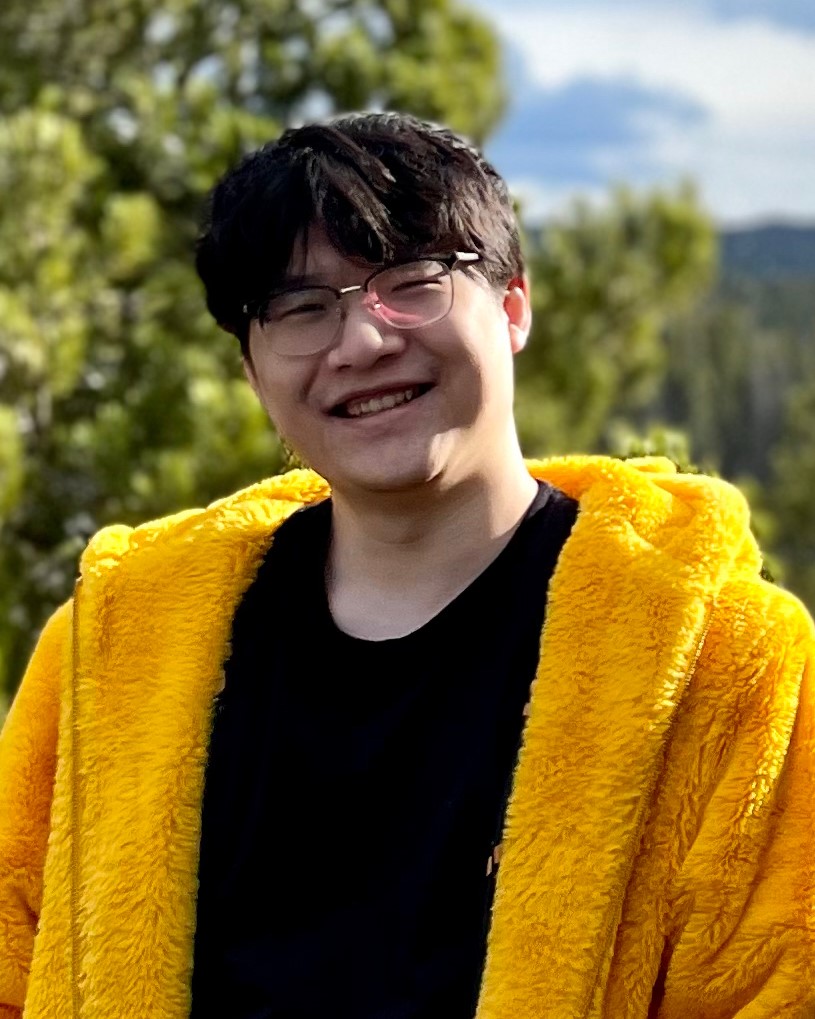}}]
{Liangqi Yuan}
(Student Member, IEEE) received the B.E. degree in Photo-electronic Information Science and Engineering from Beijing Information Science and Technology University, Beijing, China, in 2020, and the M.S. degree in Electrical and Computer Engineering from Oakland University, Rochester, MI, USA, in 2022. He is currently pursuing the Ph.D. degree in Electrical and Computer Engineering in the School of Electrical and Computer Engineering at Purdue University, West Lafayette, IN, USA. His research interests include multimodal learning, mobile computing, and machine learning.
\end{IEEEbiography}

\vspace{-10mm}

\begin{IEEEbiography}
[{\includegraphics[width=1in,height=1.25in,clip,keepaspectratio]{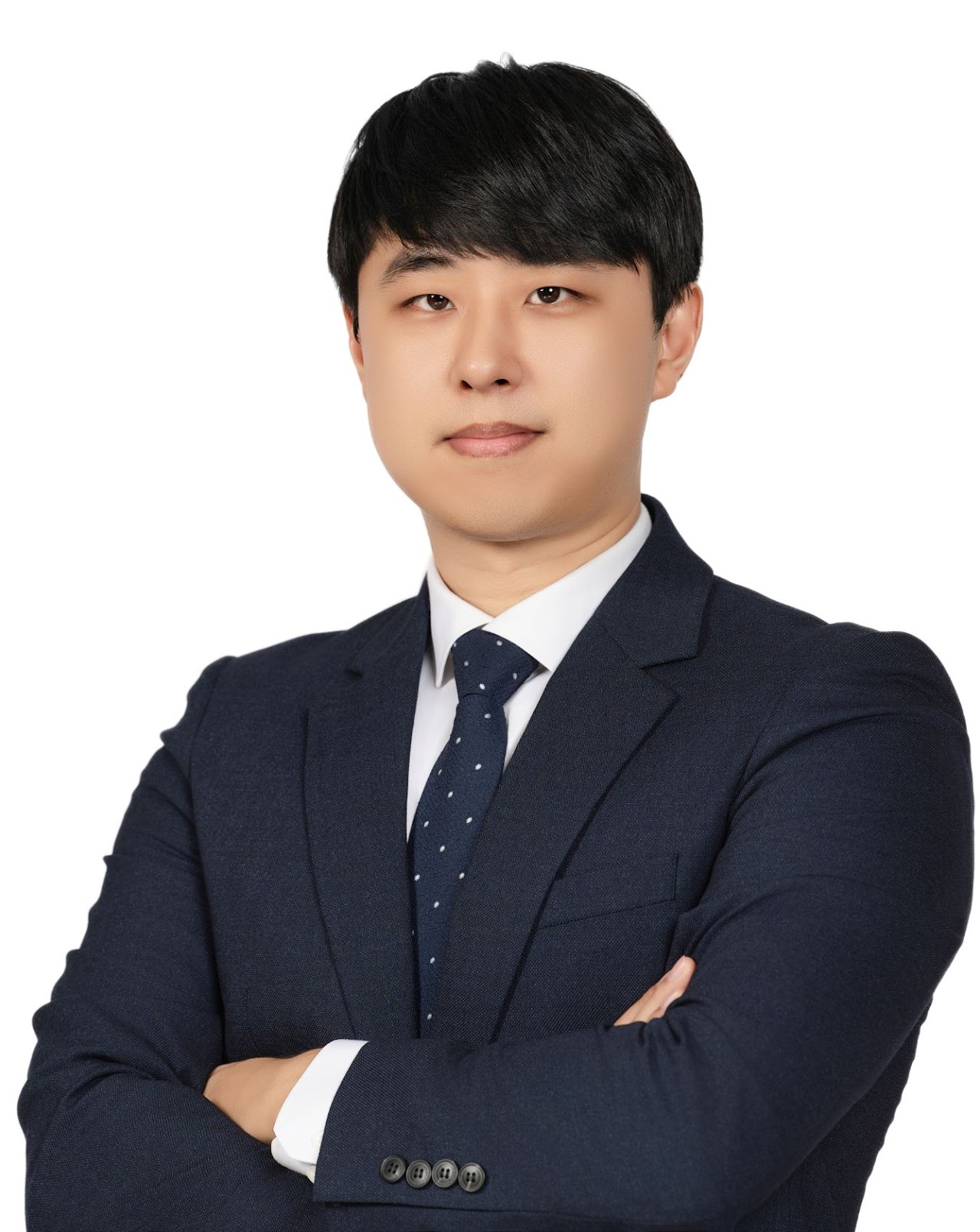}}] {Dong-Jun Han} 
(Member, IEEE) received the B.S. degree in mathematics and electrical engineering and the M.S. and Ph.D. degrees in electrical engineering from Korea Advanced Institute of Science and Technology (KAIST), South Korea, in 2016, 2018, and 2022, respectively. He is currently an Assistant Professor with the Department of Computer Science and Engineering, Yonsei University, South Korea. His research interests include the intersection of communications, networking, and machine learning, specifically in distributed/federated machine learning and network optimization.
\end{IEEEbiography}

\vspace{-10mm}

\begin{IEEEbiography}[{\includegraphics[width=1in,height=1.25in,clip,keepaspectratio]{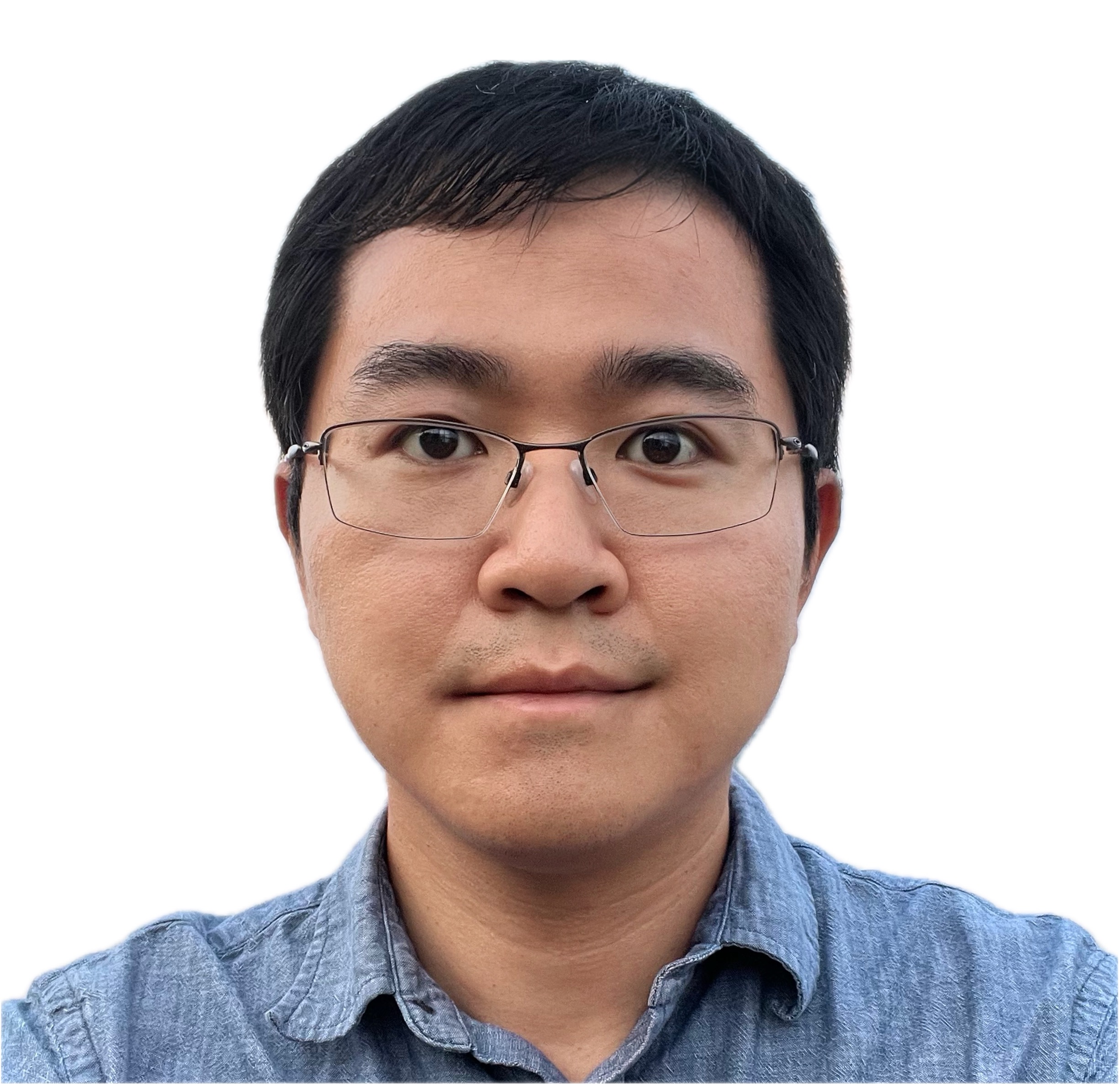}}]{Shiqiang Wang}
(Fellow, IEEE) is a Professor of Artificial Intelligence in the Department of Computer Science, University of Exeter, United Kingdom. He was a researcher at IBM T. J. Watson Research Center, NY, United States until Oct. 2025. He received his Ph.D. from Imperial College London, United Kingdom, in 2015. His research focuses on the intersection of artificial intelligence (AI), distributed computing, and optimization, with a broad range of applications including large language models (LLMs), agentic AI, efficient model training and inference, and AI in distributed systems. 
He received the IEEE Communications Society (ComSoc) Leonard G. Abraham Prize in 2021, IEEE ComSoc Best Young Professional Award in Industry in 2021, Best Paper Runner-Up of ACM MobiHoc 2025, IBM Outstanding Technical Achievement Awards (OTAA) in 2019, 2021, 2022, and 2023, and multiple Invention Achievement Awards from IBM since 2016. He is an IEEE Fellow in the Class of 2026.
\end{IEEEbiography}

\vspace{-10mm}

\begin{IEEEbiography}
[{\includegraphics[width=1in,height=1.25in,clip,keepaspectratio]{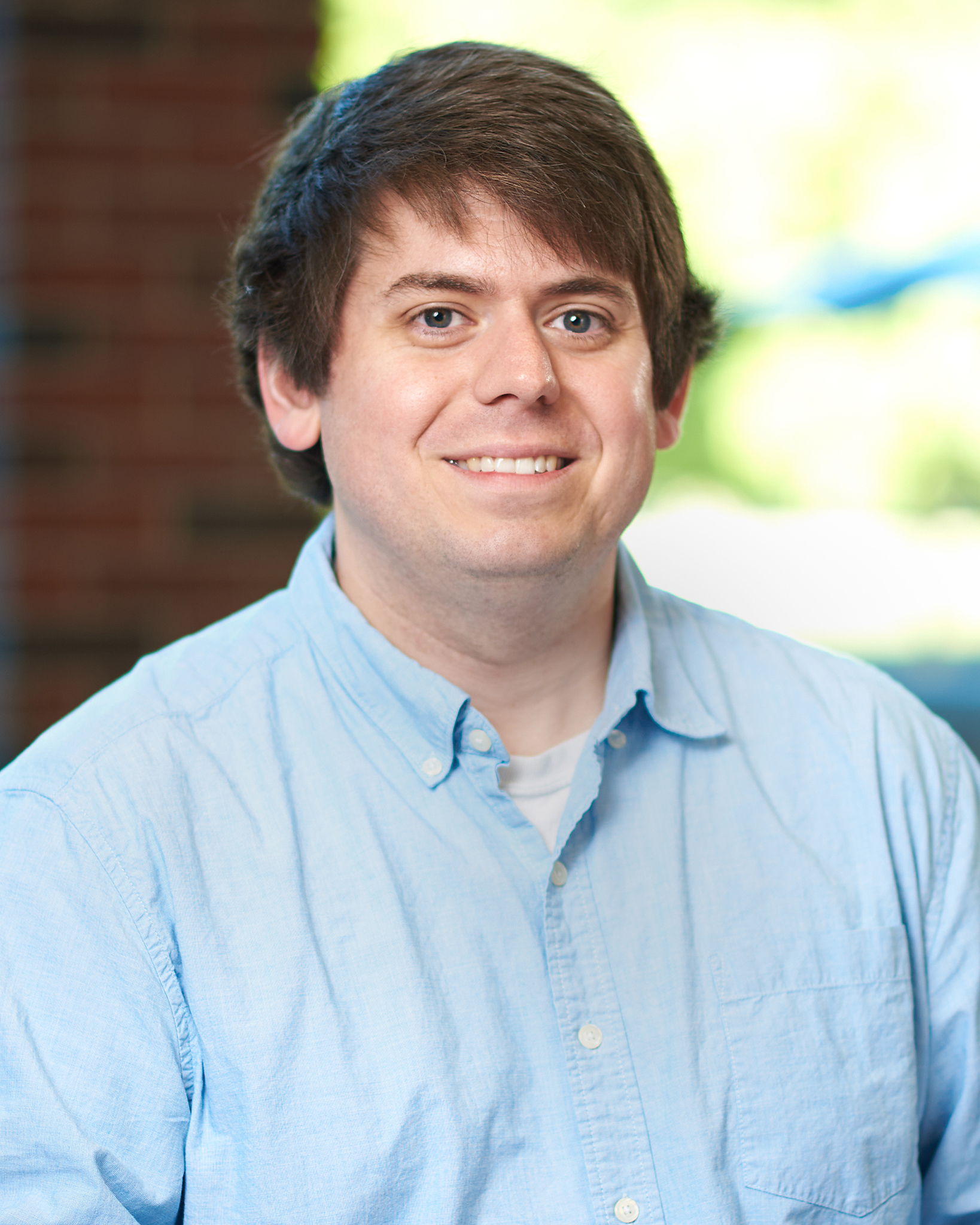}}]
{Christopher G. Brinton}
(Senior Member, IEEE) is the Elmore Associate Professor of ECE at Purdue University. He received his Ph.D. and M.Sc. in EE from Princeton University in 2016 and 2013, respectively. He is a recipient of four of the US top early career awards, from the National Science Foundation (CAREER), Office of Naval Research (YIP), Defense Advanced Research Projects Agency (YFA), and Air Force Office of Scientific Research (YIP). He is also a recipient of the Intel Rising Star Faculty Award and the Qualcomm Faculty Award.
\end{IEEEbiography}

\end{document}